\pdfoutput=1

\documentclass[11pt]{article}

\usepackage[table]{xcolor}

\usepackage[]{ACL2023}

\usepackage{times}
\usepackage{latexsym}

\usepackage[T1]{fontenc}

\usepackage[utf8]{inputenc}

\usepackage{microtype}

\usepackage{inconsolata}
\usepackage{graphicx}
\usepackage[most]{tcolorbox}
\usepackage{multirow}
\usepackage{algorithm}
\usepackage{algpseudocode}
\usepackage{amsmath}
\usepackage{todonotes}
\usepackage{amssymb}
\usepackage{pifont}
\usepackage{subcaption}
\usepackage{tabularx}

\iftrue

\newcommand{\SideNote}[2]{\todo[color=#1,size=\small]{#2}} 
\newcommand{\xueqing}[1]{\SideNote{purple!40}{#1 - Xueqing}}

\newcommand{\telins}[1]{\SideNote{green!40}{#1 -Telin}}

\else

\newcommand{\xueqing}[1]{}
\newcommand{\telins}[1]{}

\fi

\newcommand{\mypar}[1]{\vspace{.3em}\noindent\textbf{#1}}
\newcommand{\red}[1]{{\color{red} #1}}
\definecolor{mygreen}{HTML}{00A64F}
\newcommand{\green}[1]{{\color{mygreen} #1}}

\newcommand{\methodname}[0]{VDebugger}
\newcommand{\correct}[0]{\text{\ding{51}}}
\newcommand{\inc}[0]{\text{\ding{55}}}

\title{\raisebox{-0.6ex}{\includegraphics[height=1.2em]{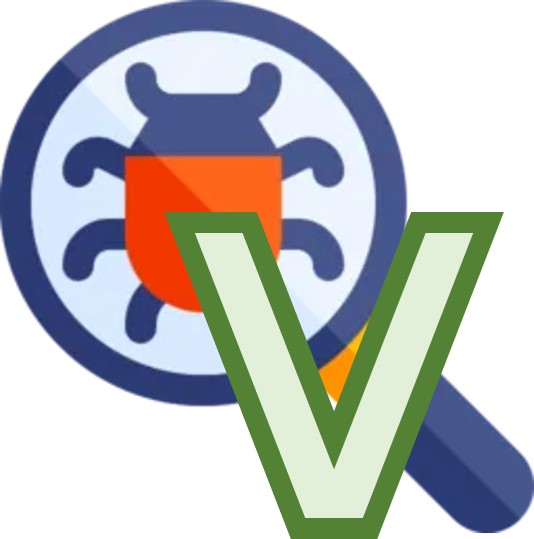}} \methodname{}: Harnessing Execution Feedback for Debugging Visual Programs}

\author{Xueqing Wu, Zongyu Lin, Songyan Zhao, Te-Lin Wu, Pan Lu, \\
\textbf{Nanyun Peng, Kai-Wei Chang} \\
University of California, Los-Angeles \\
\texttt{\{xueqing.wu,linzongy21,songyan,telinwu,pan.lu,violetpeng,kwchang\}@cs.ucla.edu} \\
\url{https://shirley-wu.github.io/vdebugger/index.html}}

\begin{document}
\maketitle
\begin{abstract}
Visual programs are executable code generated by large language models to address visual reasoning problems. They decompose complex questions into multiple reasoning steps and invoke specialized models for each step to solve the problems. However, these programs are prone to logic errors, with our preliminary evaluation showing that 58\% of the total errors are caused by program logic errors. Debugging complex visual programs remains a major bottleneck for visual reasoning. To address this, we introduce \textbf{\methodname{}}, a novel critic-refiner framework trained to localize and debug visual programs by tracking execution step by step. \methodname{} identifies and corrects program errors leveraging detailed execution feedback, improving interpretability and accuracy. The training data is generated through an automated pipeline that injects errors into correct visual programs using a novel mask-best decoding technique. Evaluations on six datasets demonstrate \methodname{}'s effectiveness, showing performance improvements of up to 3.2\% in downstream task accuracy. Further studies show \methodname{}'s ability to generalize to unseen tasks, bringing a notable improvement of 2.3\% on the unseen COVR task. Code, data and models are made publicly available at \url{https://github.com/shirley-wu/vdebugger/}.

\end{abstract}

\begin{figure*}[!t]
    \centering
\includegraphics[width=\linewidth]{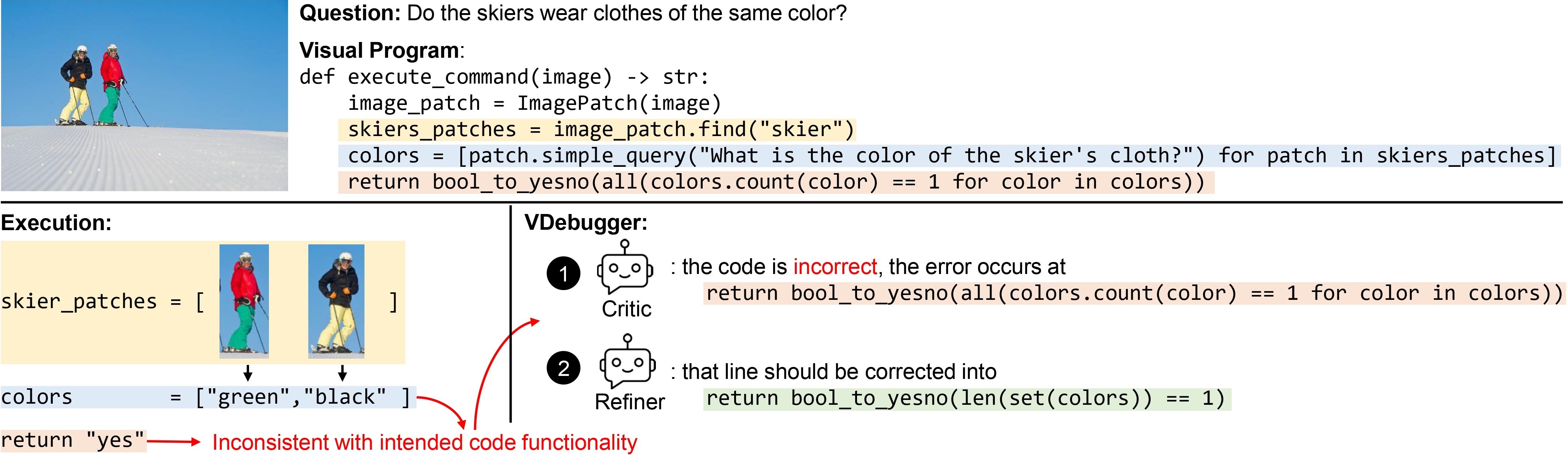}
    \caption{\textbf{Overview of visual programming and \methodname{}.} \textbf{Above}: the \textit{visual program} invokes APIs to answer the input question. Each involved API (e.g. \texttt{find}) is implemented with a specialized foundation VLM (e.g. object detection model). \textbf{Below:} \methodname{} debugs the visual program by inspecting the execution process. In this example, the \texttt{colors} variable represents the colors of all skier's jackets and contains two values, but the return value \texttt{"yes"} suggests that all skiers wear jackets of the same color. Catching this discrepancy, the critique identifies that the last line of the program is incorrect, and the refiner rewrites that line into the correct code.}
    \label{fig:teaser}
\end{figure*}

\section{Introduction}

Complex visual reasoning is a crucial yet challenging problem that often requires compositionally synthesizing multiple reasoning steps before drawing the final conclusion. For example, to answer the visual question in Figure \ref{fig:teaser}: \textit{``Do the skiers wear jackets of the same color?''}, one must identify all skiers, determine the colors of their jackets, and assess whether the colors are the same. End-to-end vision-language models (VLMs) excel at individual tasks such as object detection \citep{glip} and visual instruction following \citep{llava}. However, they struggle to generalize to complex tasks requiring compositional reasoning and inherently lack interpretability \citep{vipergpt,DBLP:conf/iclr/Yuksekgonul0KJ023,kamath2023s,kamath2024hard}.

To devise a more interpretable and generalizable reasoning process, a recent approach leverages the code generation capabilities of large language models (LLMs) to generate ``\textit{visual programs}'' \citep{vipergpt,visual_programming}. As shown in Figure \ref{fig:teaser}, the visual program decomposes a complex question into a sequence of programmatically executable steps. During execution, the visual program invokes foundational specialist models to perform visual perception and synthesize the results of each reasoning step into the final answer. The inherent compositionality of programs allows this approach to perform compositional reasoning while ensuring generalization and interpretability.

Nonetheless, program errors become a bottleneck for this approach, accounting for 58\% of total errors as shown in our evaluation. Following the advancement of LLM self-refinement in general-domain code generation \citep{selfdebugging} and LLM agents \citep{selfrefine,reflexion}, recent work leverages zero-shot prompting of LLMs to debug visual programs based on some given feedback \citep{towards_zero_shot_visual_programming,define}. However, their feedback typically focuses on limited aspects such as compilation errors. Furthermore, the zero-shot prompting technique is less effective for self-critique and self-correction of programs, especially for smaller LLMs, as shown in recent work \cite{luo2023critique,tian2024debugbench,lan2024criticbench,jiang2024training}.

In this work, we propose \textbf{\methodname{}}, a tool trained to debug visual programs by tracking their execution step by step. As shown in Figure \ref{fig:teaser}, \methodname{} takes as input the execution states at each step, including the code being executed and the resulting change of variable values. Based on such information, the \textbf{critic} identifies fine-grain program errors down to the line, and the \textbf{refiner} rewrites the error-inducing line to correct the program.

To train the \methodname{}, we devise an automated pipeline to collect training data at scale. For visual question answering task, speficially, we prompt an LLM to generate visual programs for the input questions from existing datasets. The programs whose execution results match the ground truth answers are taken as correct programs. In order to create incorrect programs, we inject errors by resampling parts of these originally correct programs and thereby generating modifications that affect the execution results. The \methodname{} thus learns to identify and correct visual program errors utilizing these automatically curated positive and negative program pairs. In particular, we propose a \textit{mask-best sampling} algorithm that increases the success rate of error injection by up to 10 times compared to greedy decoding. Eventually, we generate a total of $47.7k$ program pairs for \methodname{} training.

We evaluate \methodname{} on a total of 6 datasets covering various forms of visual question answering \citep{gqa,tallyqa,suhr-etal-2019-corpus} and visual grounding \citep{refcoco}. Based on both CodeLlama-7B and CodeLlama-13B, \methodname{} consistently improves the performance by up to $3.2\%$ accuracy. \methodname{} can also be employed to debug visual programs generated by proprietary code generation models such as GPT-3.5 and brings notable gains of up to $4.9\%$ accuracy. By jointly training \methodname{} on all six datasets with different task forms, \methodname{} demonstrates generalization capability capable of handling unseen tasks such as question answering based on variable number of images \citep{bogin-etal-2021-covr}.

In summary, our contributions are three-folds: (1) We propose \methodname{}, a novel framework for debugging visual programs capable of reasoning over execution process and performing explainable debugging; (2) We develop a pipeline to automatically generate large-scale training datasets including $47.7k$ program pairs; (3) Our \methodname{} trained on top of 7B and 13B LLMs achieves significant improvements across 6 datasets and can generalize to unseen scenarios.

\section{Related Work}

\begin{table*}[!htbp]
\small
    \centering
    \setlength\tabcolsep{5.5pt}
    \begin{tabular}{l|l|ccc|c}
    \hline
    & \multirow{2}*{Method} & \multicolumn{3}{c|}{Feedback} & \multirow{2}*{Training} \\
    & & Error & Unit-test & Execution states \\
    \hline
    \multirow{3}*{LLM self-refinement} & SelfRefine \citep{selfrefine} & N/A & N/A & \red{\inc{}} & \red{\inc{}} \\
    & Reflexion \citep{reflexion} & N/A & N/A & \red{\inc{}} & \red{\inc{}} \\
    & Refiner \citep{paul-etal-2024-refiner} & N/A & N/A & Per step & \green{\correct{}} \\
    \hline
    \multirow{3}*{General-domain code debugging} & SelfDebug \citep{selfdebugging} & \green{\correct{}} & \green{\correct{}} & \red{\inc{}} & \red{\inc{}} \\
    & \citet{jiang2024training} & \green{\correct{}} & \green{\correct{}} & \red{\inc{}} & \green{\correct{}} \\
    & LDB \citep{ldb} & \green{\correct{}} & \green{\correct{}} & Per block & \red{\inc{}} \\
    \hline
    \multirow{3}*{Visual program debugging} & \citet{towards_zero_shot_visual_programming} & \green{\correct{}} & N/A & \red{\inc{}} & \red{\inc{}} \\
    & Define \citep{define} & \green{\correct{}} & N/A & Per block & \red{\inc{}} \\
    & \textbf{\methodname{} (Ours)} & \green{\correct{}} & N/A & Per step & \green{\correct{}} \\
    \hline
    \end{tabular}
    \caption{\textbf{Comparison against existing work.} The distinction of our \methodname{} against existing work are mainly two-folds: (1) we utilize a more fine-grained feedback information of step-wise execution states, and (2) we automatically collect large-scale training data for model training.}
    \label{tab:baseline}
\end{table*}

\mypar{Visual reasoning.} The large-scale pre-training of VLMs has demonstrated significant success \citep{clip}. When fine-tuned, these models can effectively adapt to specific tasks such as instruction following \citep{llava,qwenvl}, visual question answering \citep{blip,blip2}, and object detection \citep{glip}. Despite their impressive performance on these individual tasks, VLMs still struggle with compositional reasoning that requires composing multiple reasoning steps \citep{gqa,suhr-etal-2019-corpus,bogin-etal-2021-covr}.
Visual programming addresses this problem \citep{vipergpt,visual_programming} by generating executable programs that decompose the question into multiple reasoning steps and invoke specialized VLMs for each step. However, the program errors in the generated code become a bottleneck of this approach.

\mypar{Self-debugging and self-refinement.} We present a comprehensive comparison between this work and existing work for self-debugging and self-refinement in Table \ref{tab:baseline}. Existing techniques have explored LLM self-refinement for reasoning, decision making, and language generation tasks \citep{selfrefine,reflexion}. These work largely relies on self-generated feedback, which is less effective especially for code-related tasks \citep{DBLP:journals/corr/abs-2310-01798}. \citet{paul-etal-2024-refiner} tracks the intermediate states step-by-step during mathematical reasoning, which is shown to be more beneficial.
For general-domain code generation, self-debugging can leverage more reliable feedback information such as execution error and pass/fail results of unit-tests \citep{selfdebugging}. \citet{ldb} further divides a program into multiple code blocks and takes the execution states before and after each block as feedback. However, visual programs do not have unit-tests available. Existing work for debugging visual programs either use execution error \citep{towards_zero_shot_visual_programming} or block-wise execution states \citep{define} as feedback, which may not be fine-grained enough to cover all potential errors. Our feedback is more informative by tracking execution states  step-by-step. Another trend of recent work is to generate synthetic data for training self-debugging models, which is particularly helpful for smaller LLMs \citep{paul-etal-2024-refiner,jiang2024training}. We follow this trend to collect large-scale training sets for training \methodname{}.

\begin{algorithm}
\small
\caption{\methodname{} algorithm}
\label{alg:pipeline}
\begin{algorithmic}[1] %
\Require Critic $C$, refiner $R$, score threshold $th$, max step $T$, initial program $P_0$
\State $P = P_0$
\For{$i = 1$ \textbf{to} $T$}
    \State $fb = \textsc{Execute}(P)$ \Comment{Collect feedback}
    \State $score, loc = C(P, fb)$ \Comment{Identify and localize error}
    \If{$score > th$}  \Comment{Correct program}
        \State \Return $P$
    \EndIf
    \State $P_{new} = R(P, fb, loc)$ \Comment{Refine program}
    \State $P = P_{new}$
\EndFor
\State \Return $P$
\end{algorithmic}
\end{algorithm}

\section{\methodname{} Framework}

\methodname{} consists of two components, a \textbf{critic} and a \textbf{refiner}.
The debugging process is illustrated in Alg. \ref{alg:pipeline}. 
Starting with an initial program $P_0$ and its execution feedback, the critic model $C$ detects and localizes potential errors. Subsequently, the refiner $R$ corrects these identified errors. This iterative process continues until the critic model $C$ deems the program satisfactory.\footnote{While this is a general framework applicable to various programming languages, this work focuses on Python.}

\begin{figure*}[!htbp]
    \centering
\includegraphics[width=\linewidth]{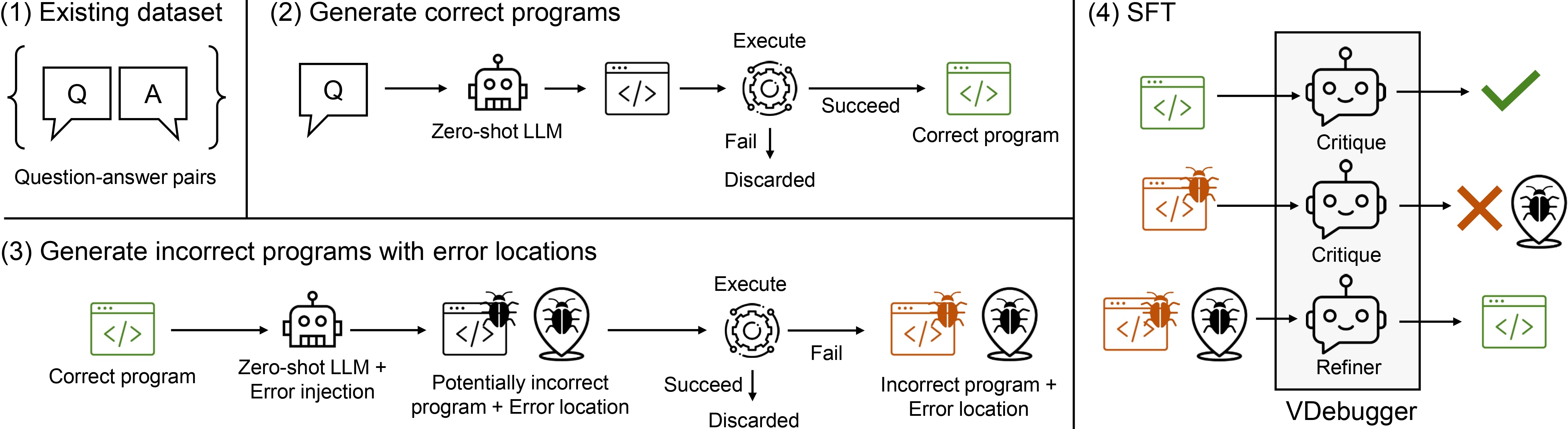}
    \caption{\textbf{Training data collection pipeline.} Given an existing dataset of question-anwswer pairs, we prompt LLM to generate correct programs, inject error to generate incorrect programs, and use the paired data for SFT training.}
    \label{fig:data}
\end{figure*}

\mypar{Execution feedback.} In contrast to previous approaches that focus on execution errors and block-wise execution states \citep{towards_zero_shot_visual_programming,define}, our objective is to develop a general and comprehensive feedback mechanism that can cover a wider range of errors. Drawing inspiration from the stepping debugging strategy\footnote{\url{https://en.wikipedia.org/wiki/Stepping_(debugging)}} of human programmers, 
we track the execution process step by step and document each executed program line, the resulting changes in the intermediate variables, and any errors encountered during execution. This feedback information is fed to \methodname{} in text format as in Figure \ref{fig:trace} in the Appendix.

\mypar{Critic.} Critic $C$ jointly detects and localizes the error in the program. Formally, given the input program $P$ and its feedback information collected through execution (denoted as $\textsc{Execute}(P)$),
\begin{align*}
    score, loc = C(P, \textsc{Execute}(P)),\label{eq:critic_scorer}
\end{align*}
where $score$ represents how likely the program $P$ is correct. $P$ is considered correct when $score$ exceeds a threshold $th$ (0.5 in this work). The critic classify the program $P$ into either correct or incorrect. Concretely, $C$ first generates a correctness token chosen from $\{t_\correct{}, t_\inc{}\}$ representing whether the program is correct or incorrect, so the probability assigned to token $t_\correct{}$ can serve as $score$. If the token $t_\inc{}$ is generated, $C$ further generates the error location $loc$. Here, the location $loc$ is a span within program P defined by its start and end positions, which can be a word, a line, multiple lines, or any continuous segment.

\mypar{Refiner.} Conditioned on the error location $loc$, refiner $R$ rewrites location $loc$ to fix the program. Formally,
\begin{align*}
& P_{new} = R(P, \textsc{Execute}(P), loc),
\end{align*}
where the output program $P_{new}$ only differs with the input program $P$ at location $loc$.

\section{Training of \methodname{}}

\begin{table*}
    \centering
    \small
    \begin{tabular}{l|rr|rr|rr|r}
    \hline
    & \multirow{2}*{$|\mathcal{P}^{(0)}_{\inc{}}|$} & \multirow{2}*{$|\mathcal{P}^{(0)}_{\correct{}}|$} & \multicolumn{2}{c|}{Greedy} & \multicolumn{2}{c|}{Mask-best} & \multirow{2}*{Final} \\
    & & & $|\mathcal{P}^{(1)}_{\inc{}}|$ & Error Rate & $|\mathcal{P}^{(1)}_{\inc{}}|$ & Error Rate\\
    \hline
    GQA & 21,874 & 18,126 & 3,927 & 21.7\% & 7,758 & 42.8\% & 11,188 \\
    TallyQA & 17,310 & 22,690 & 842 & 3.7\% & 8,843 & 38.9\% & 9,593 \\
    NLVRv2 & 19,415 & 25,085 & 3,803 & 15.2\% & 10,263 & 40.9\% & 13,948 \\
    RefCOCO & 21,013 & 18,987 & 4,710 & 24.8\% & 8,635 & 45.5\% & 12,949 \\
    \hline
    \end{tabular}
    \caption{\textbf{Statistics of collected training data.} We report the number of correct and incorrect programs in the initial pool (denoted as $|\mathcal{P}^{(0)}_{\inc{}}|$ and $|\mathcal{P}^{(0)}_{\correct{}}|$), the number of incorrect programs generated via greedy decoding and mask-best decoding (denoted as $|\mathcal{P}^{(1)}_{\inc{}}|$), and the rate at which an error is successfully injected computed as $|\mathcal{P}^{(1)}_{\inc{}}|/|\mathcal{P}^{(0)}_{\correct{}}|$ (denoted as Error Rate). In total, we collect 47,678 paired training data.}
    \label{tab:datasets}
\end{table*}

With the critic-refiner framework introduced above, now we design an automated pipeline to collect training data tailored for our framework.
Our goal is to obtain tuples $\{(P_\correct{}, P_\inc{}, loc)\}$ where in each tuple, the correct program $P_\correct{}$ and incorrect one $P_\inc{}$ only differ at location $loc$. As in Figure \ref{fig:data}, our pipeline consists of two steps: (1) generating correct programs, and (2) generating incorrect programs with error locations.

\mypar{Correct program generation.}
Given pairs of questions and ground truth answers from existing datasets, we prompt LLM to generate an initial pool of visual programs denoted as $\mathcal{P}^{(0)}$. The subset of programs whose execution results match the ground truth labels (denoted as $\mathcal{P}^{(0)}_{\correct{}}$) will be kept for the next step, while the rest of the programs (denoted as $\mathcal{P}^{(0)}_{\inc{}}$) will be discarded. %

\mypar{Incorrect program generation.} For each correct program $P \in \mathcal{P}^{(0)}_{\correct{}}$, we obtain a potentially incorrect program $P'$ by resampling part of the program $P$ at a random location $loc$. We then execute program $P'$ and select those whose execution results do not match ground truth labels, denoted as $\mathcal{P}^{(1)}_{\inc{}}$. %

\begin{figure}[!t]
    \centering
\includegraphics[width=0.9\linewidth]{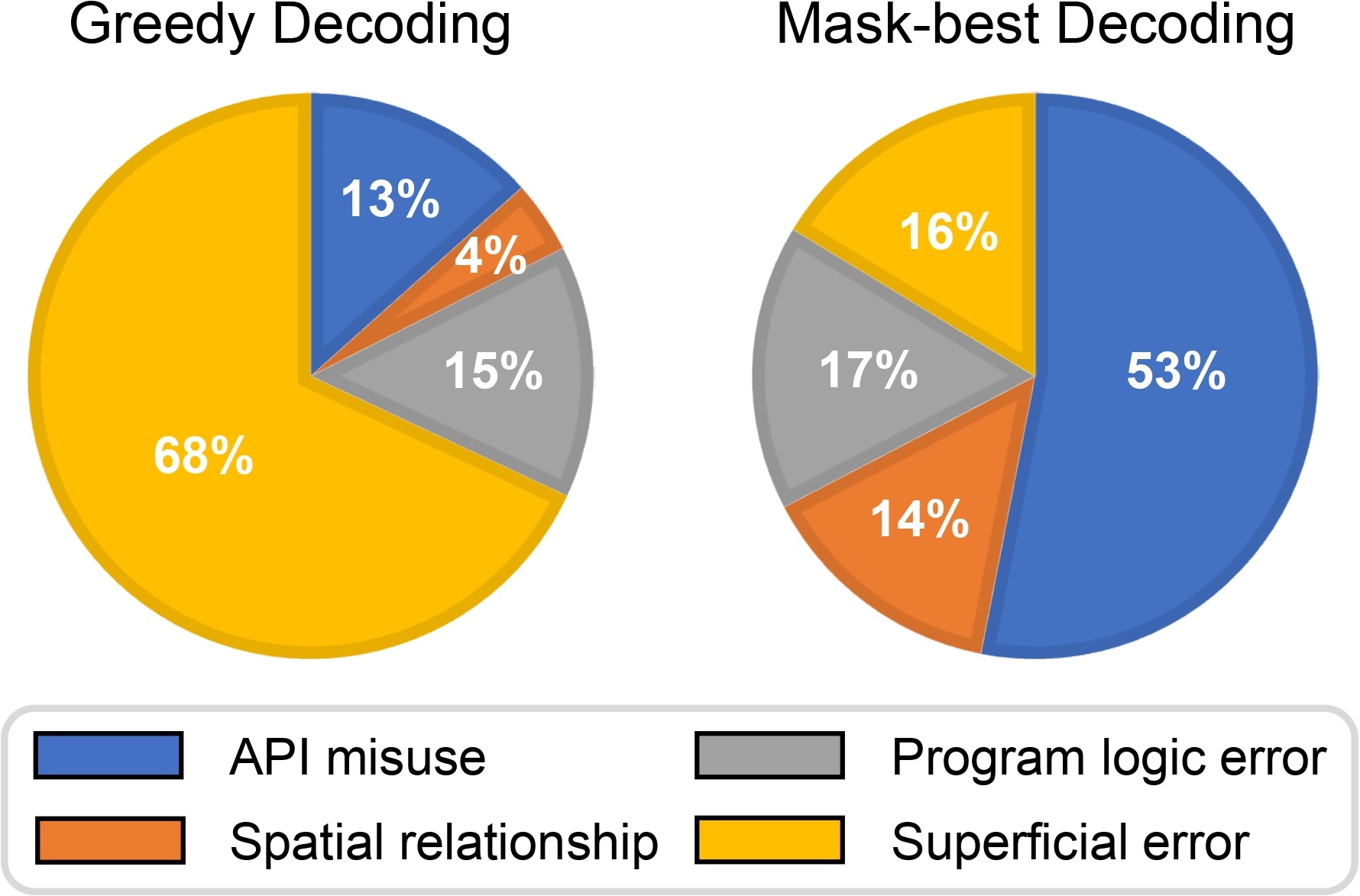}
    \caption{\small Categorization of synthetic errors generated by greedy decoding and mask-best decoding respectively.}
    \label{fig:synthetic_error}
\end{figure}

Concretely, we first parse the correct program $P$ into a abstract syntax tree and randomly sample a subtree as location $loc$. We then mask out the selected location and prompt LLM to recover the masked content. To more effectively inject errors to the location, we propose a \textbf{mask-best sampling} strategy. At each decoding step, given the probability distribution $p$ predicted by LLM, we mask out the token $i^*$ with highest probability and only sample from the tail distribution $p^{(tail)}$:
\begin{align*}
    &i^* = \arg \max_i p_i \\
    &p^{(tail)}_i = \begin{cases}
        0 & i = i^* \\
        p_i / \left(1 - p_{i^*}\right) & i \neq i^*.
    \end{cases}
\end{align*}
To ensure output quality, we only apply mask-best sampling to tokens with low confidence, determined as follows:
\begin{align*}
    & p_{i^*} - p_{i^{(2)}} < th,\ i^{(2)} = \arg \max_{i \neq i^*} p_i,
\end{align*}
and we apply mask-best sampling to at most $N$ tokens. The threshold $th$ is set as 0.9 in this work. The formal algorithm is in Alg. \ref{alg:sampling}. As shown in Table \ref{tab:datasets}, mask-best dramatically increases the rate at which an error is successfully injected by up to 10 times (from $3.7\%$ to $38.9\%$). We manually analyze and categorize 200 errors injected into GQA dataset. As shown in Figure \ref{fig:synthetic_error}, greedy sampling generates a large number of superficial errors referencing variables before their creation. In contrast, mask-best sampling produces a broader range of more complex and diverse errors.

\begin{algorithm}
\small
\caption{Mask-best sampling}
\label{alg:sampling}
\begin{algorithmic}[1]
\Require LLM, prompt, confidence threshold $th$, maximum numbers for mask-best sampling $N$, maximum number of tokens $T$
\State $P = []$ \Comment{Empty string for sampling}
\State $n = 0$ \Comment{Mask-best sampling counter}
\For{$i = 1$ \textbf{to} $T$}
    \State $p = \text{LLM}(\text{prompt}, P)$
    \If{$n < N$ and $p_{i^*} - p_{i^{(2)}} < th$}
        \State $p = p^{(tail)}$ \Comment{Sample from the tail}
        \State $n = n + 1$
    \EndIf
    \State $P = [P ; \textsc{Sample}(p)]$
    \If{EOS is sampled}
        \State break
    \EndIf
\EndFor
\State \Return $P$
\end{algorithmic}
\end{algorithm}

\mypar{Training.} Pairing programs from $\mathcal{P}_{\correct{}}^{(0)}$ and $\mathcal{P}_{\inc{}}^{(1)}$, we obtain a training set $\{(P_\correct{}, P_\inc{}, loc)\}$ for training the critic $C$ and refiner $R$. Our training objectives are as follows:
\begin{align*}
\mathcal{L}_{C} &= \sum \mathcal{L}\left(t_\inc{}, loc| P_\inc{}, \textsc{Execute}(P_\inc{})\right) + \\
&\qquad \mathcal{L}\left(t_\correct{}| P_\correct{}, \textsc{Execute}(P_\correct{})\right)), \\
\mathcal{L}_{R} &= \sum \mathcal{L}(P_\correct{}| P_\inc{}, \textsc{Execute}(P_\inc{}), loc).
\end{align*}
where $\mathcal{L}$ represents autoregressive language modeling objective.
However, training $C$ only on programs with injected errors limits its ability to detect errors in naturally generated programs due to the distribution shift. Leveraging the large pool of incorrect programs $\mathcal{P}^{(0)}_{\inc{}}$ generated in the first step, we introduce an additional objective to address the distribution shift:
\begin{align*}
\mathcal{L}_{C'} &= \sum_{P_\correct{}\in \mathcal{P}^{(0)}_{\correct{}}} \mathcal{L}(t_\correct{}|P_\correct{}, \textsc{Execute}(P_\correct{})) + \\
&\quad \sum_{P_\inc{}\in \mathcal{P}^{(0)}_{\inc{}}} \mathcal{L}(t_\inc{}|P_\inc{}, \textsc{Execute}(P_\inc{})),
\end{align*}
and the final training objective for $C$ is $\mathcal{L}_{C,final} = \mathcal{L}_C + \mathcal{L}_{C'}$.
The mixed objective enables $C$ to detect and localize errors in naturally generated programs without requiring error location annotations for these programs.

\definecolor{highlight_bg}{rgb} {0.88, 0.94, 1.0} %

\begin{table*}[!htb]
\centering
\setlength{\tabcolsep}{2pt}
\resizebox{\textwidth}{!}{
\begin{tabular}{l|llllll|l}
\hline
& GQA & TallyQA & NLVRv2 & RefCOCO & RefCOCO+ & RefCOCOg & Mean \\
\hline
Base VLM & 44.7 & 29.9 & 55.4 & 55.0 & 52.2 & 50.1 & 47.9 \\
\hline
\multicolumn{5}{l}{\textit{CodeLlama-7B as code generator}} \\
\hline
No Debugging \citep{vipergpt} & 43.1 & 45.4 & 59.7 & 56.2 & 51.6 & 53.9 & 51.7 \\
SelfDebug \citep{selfdebugging} & 42.7 & 44.4 & 61.4 & 55.9 & 51.4 & 52.3 & 51.4 \\
LDB \citep{ldb} & 41.8 & 39.4 & 56.4 & 50.4 & 51.8 & 52.2 & 48.7 \\
\rowcolor{highlight_bg} \textbf{\methodname{}} (Ours) & \textbf{46.3} {\footnotesize \color{mygreen} (+3.2)} & \textbf{46.4} {\footnotesize \color{mygreen} (+1.0)} & 61.4 {\footnotesize \color{mygreen} (+1.7)} & \textbf{58.7} {\footnotesize \color{mygreen} (+2.5)} & \textbf{52.3} {\footnotesize \color{mygreen} (+0.7)} & \textbf{56.3} {\footnotesize \color{mygreen} (+2.4)} & \textbf{53.6} {\footnotesize \color{mygreen} (+1.9)} \\
\rowcolor{highlight_bg} \textbf{\methodname{} w/ Gen} (Ours) & 46.0 {\footnotesize \color{mygreen} (+2.9)} & 46.3 {\footnotesize \color{mygreen} (+0.9)} & \textbf{61.8} {\footnotesize \color{mygreen} (+2.1)} & 58.3 {\footnotesize \color{mygreen} (+2.1)} & 51.8 {\footnotesize \color{mygreen} (+0.2)} & 55.9 {\footnotesize \color{mygreen} (+2.0)} & 53.3 {\footnotesize \color{mygreen} (+1.6)} \\
\hline
\multicolumn{5}{l}{\textit{CodeLlama-13B as code generator}} \\
\hline
No Debugging \citep{vipergpt} & 45.4 & 47.7 & 64.8 & 56.7 & 54.7 & 55.1 & 54.1 \\
SelfDebug \citep{selfdebugging} & 41.6 & 42.8 & 62.5 & 41.2 & 39.2 & 46.2 & 45.6 \\
LDB \citep{ldb} & 42.7 & 37.4 & 57.1 & 51.0 & 51.1 &  50.8 & 48.3 \\
\rowcolor{highlight_bg} \textbf{\methodname{}} (Ours) & \textbf{48.1} {\footnotesize \color{mygreen} (+2.7)} & 48.3 {\footnotesize \color{mygreen} (+0.6)} & 65.0 {\footnotesize \color{mygreen} (+0.2)} & 58.1 {\footnotesize \color{mygreen} (+1.4)} & \textbf{55.5} {\footnotesize \color{mygreen} (+0.8)} & \textbf{58.3} {\footnotesize \color{mygreen} (+3.2)} & 55.5 {\footnotesize \color{mygreen} (+1.4)} \\
\rowcolor{highlight_bg} \textbf{\methodname{} w/ Gen} (Ours) & \textbf{48.1} {\footnotesize \color{mygreen} (+2.7)} & \textbf{48.6} {\footnotesize \color{mygreen} (+0.9)} & \textbf{65.8} {\footnotesize \color{mygreen} (+1.0)} & \textbf{58.5} {\footnotesize \color{mygreen} (+1.0)} & 55.0 {\footnotesize \color{mygreen} (+0.3)} & 58.2 {\footnotesize \color{mygreen} (+3.1)} & \textbf{55.7} {\footnotesize \color{mygreen} (+1.6)} \\
\hline
\end{tabular}
}
\caption{\textbf{Main results.} We report accuracy for GQA, TallyQA, NLVRv2, and IoU for RefCOCO\* datasets. We compare the performance of two debugging baselines and our \methodname{} (highlighted in the table). Here, \methodname{} w/ Gen denotes the generalist model trained on all datasets. For comparison, we also report the performance of the base VLMs.}
\label{tab:main_results}
\vspace{-1em}
\end{table*}

\section{Experiments}

In this section, we aim to: (1) evaluate the effectiveness of \methodname{} by comparing against existing self-debugging methods; (2) analyze the benefits brought by each individual component; and (3) demonstrate its generalization capability by debugging programs generated by other LLMs and by evaluating on unseen tasks.

\mypar{Dataset.} We experiment on three forms of tasks including 6 datasets: (1) \textit{Visual question answering with one image}, including GQA dataset \citep{gqa} targeting compositional question answering and TallyQA  dataset \citep{tallyqa} targeting counting; (1) \textit{Visual question answering with multiple images}, including NLVRv2 dataset \citep{suhr-etal-2019-corpus} where each question is accompanied by two images; (3) \textit{Visual grounding} including three variants of RefCOCO dataset \citep{refcoco}: the original RefCOCO dataset, RefCOCO+ that disallows location descriptions, and RefCOCOg that involves longer and more complex text descriptions. 
We report accuracy for question answering tasks and IoU for visual grounding tasks.

For training data collection, we generate 4 training sets for the GQA, TallyQA, NLVRv2 and RefCOCO datasets respectively. We use CodeLlama-7B-Python \citep{codellama} to generate the initial program pool $\mathcal{P}^{(0)}$, and use CodeLlama-7B-Instruct to generate incorrect programs $\mathcal{P}^{(1)}_{\inc{}}$ with both greedy decoding and mask-best sampling. We collect 9$\sim$14$k$ training data for each dataset and in total 47.7$k$ data. Detailed statistics are in Table \ref{tab:datasets}.

\mypar{Evaluated models.} We use ViperGPT \citep{vipergpt} as our base visual program generator before any debugging. We train \methodname{} on each dataset based on CodeLlama-7B-Python and CodeLlama-13B-Python. We further train a generalized variant on the mix of all datasets denoted as \methodname{} w/ Gen. During inference, we use a maximum iteration step of $T=3$ unless otherwise noted.
We compare our method against two code debugging methods: SelfDebug \citep{selfdebugging} and LDB \citep{ldb}. SelfDebug debugs the program based on unit-test feedback. Since visual programs do not have unit tests available, we replace it with our execution feedback. LDB uses execution states per program block to iteratively rewrite each block, making it more expensive than our strategy. Both SelfDebug and LDB relies on zero-shot prompting without any training.

\subsection{Results}
Table~\ref{tab:main_results} shows our main results on all six datasets. Both SelfDebug and LDB slightly hurt the performance, likely due to the limited self-debugging capability of small LLMs as noted by recent studies \citep{luo2023critique,tian2024debugbench,lan2024criticbench,jiang2024training}. The challenge is exacerbated by the absense of visual programs during the pre-training stage of LLMs, highlighting the necessity of training debugging models for visual programs.
In contrast, our \methodname{} consistently improves the performance in every dataset, achieving improvements of up to 3.2\% accuracy.

\begin{table*}[!htb]
    \centering
    \begin{tabular}{l|ll|ll|lll}
    \hline
& \multicolumn{2}{c|}{Critic Acc.} & \multicolumn{2}{c|}{Refiner SR} & \multicolumn{3}{c}{Task Performance} \\
Dataset & w/o FB & w/ FB & w/o FB & w/ FB & {\footnotesize No Debug} & {\footnotesize Ours w/o FB} & {\footnotesize Ours w/ FB} \\
    \hline
GQA & 69.6 & \cellcolor{mygreen!13.61}\textbf{73.9} {\color{mygreen} \footnotesize(+4.3)} & \textbf{47.7} & \cellcolor{mygreen!0.00}\textbf{47.7} {\color{mygreen} \footnotesize(+0.0)} & 43.1 & \cellcolor{mygreen!8.23}45.7 {\color{mygreen} \footnotesize(+2.6)} & \cellcolor{mygreen!10.13}\textbf{46.3} {\color{mygreen} \footnotesize(+3.2)} \\
TallyQA & 56.3 & \cellcolor{mygreen!50.00}\textbf{72.1} {\color{mygreen} \footnotesize(+15.8)} & \textbf{10.7} & \cellcolor{red!1.27}10.3 {\color{red} \footnotesize(-0.4)} & 45.4 & \cellcolor{mygreen!1.58}45.9 {\color{mygreen} \footnotesize(+0.5)} & \cellcolor{mygreen!3.16}\textbf{46.4} {\color{mygreen} \footnotesize(+1.0)} \\
NLVRv2 & 66.1 & \cellcolor{mygreen!3.80}\textbf{67.3} {\color{mygreen} \footnotesize(+1.2)} & 10.8 & \cellcolor{mygreen!13.61}\textbf{15.1} {\color{mygreen} \footnotesize(+4.3)} & 59.7 & \cellcolor{mygreen!2.22}60.4 {\color{mygreen} \footnotesize(+0.7)} & \cellcolor{mygreen!5.38}\textbf{61.4} {\color{mygreen} \footnotesize(+1.7)} \\
RefCOCO & 71.9 & \cellcolor{mygreen!28.48}\textbf{80.9} {\color{mygreen} \footnotesize(+9.0)} & 26.7 & \cellcolor{mygreen!1.58}\textbf{27.2} {\color{mygreen} \footnotesize(+0.5)} & 56.2 & \cellcolor{mygreen!6.65}58.3 {\color{mygreen} \footnotesize(+2.1)} & \cellcolor{mygreen!7.91}\textbf{58.7} {\color{mygreen} \footnotesize(+2.5)} \\
RefCOCO+ & 70.4 & \cellcolor{mygreen!22.15}\textbf{77.4} {\color{mygreen} \footnotesize(+7.0)} & 56.3 & \cellcolor{mygreen!2.53}\textbf{57.1} {\color{mygreen} \footnotesize(+0.8)} & 51.6 & \cellcolor{red!0.95}51.3 {\color{red} \footnotesize(-0.3)} & \cellcolor{mygreen!7.28}\textbf{53.9} {\color{mygreen} \footnotesize(+2.3)} \\
RefCOCOg & 64.5 & \cellcolor{mygreen!27.22}\textbf{73.1} {\color{mygreen} \footnotesize(+8.6)} & 32.2 & \cellcolor{mygreen!2.85}\textbf{33.1} {\color{mygreen} \footnotesize(+0.9)} & 53.9 & \cellcolor{mygreen!5.70}55.7 {\color{mygreen} \footnotesize(+1.8)} & \cellcolor{mygreen!7.59}\textbf{56.3} {\color{mygreen} \footnotesize(+2.4)} \\
\hline
Mean & 66.5 & \cellcolor{mygreen!24.21}74.1 {\color{mygreen} \footnotesize(+7.7)} & 30.7 & \cellcolor{mygreen!3.22}31.8 {\color{mygreen} \footnotesize(+1.0)} & 51.6 & \cellcolor{mygreen!3.90}52.9 {\color{mygreen} \footnotesize(+1.2)} & \cellcolor{mygreen!6.91}53.8 {\color{mygreen} \footnotesize(+2.2)} \\
        \hline
    \end{tabular}
    \caption{\textbf{Ablation study.} We report the critic accuracy (Acc.), the refiner success rate (SR), and final task performance on downstream tasks. For each component, we report the performance either without or with execution feedback (denoted as w/o FB and w/ FB). We also report downstream task performance before any debugging. Results are evaluated on 7B-level VDebugger models.}
    \label{tab:main_ablation}
\end{table*}

\begin{figure*}[!htb]
\centering
\small
\begin{subfigure}{.33\textwidth}
  \centering
  \includegraphics[width=\textwidth]{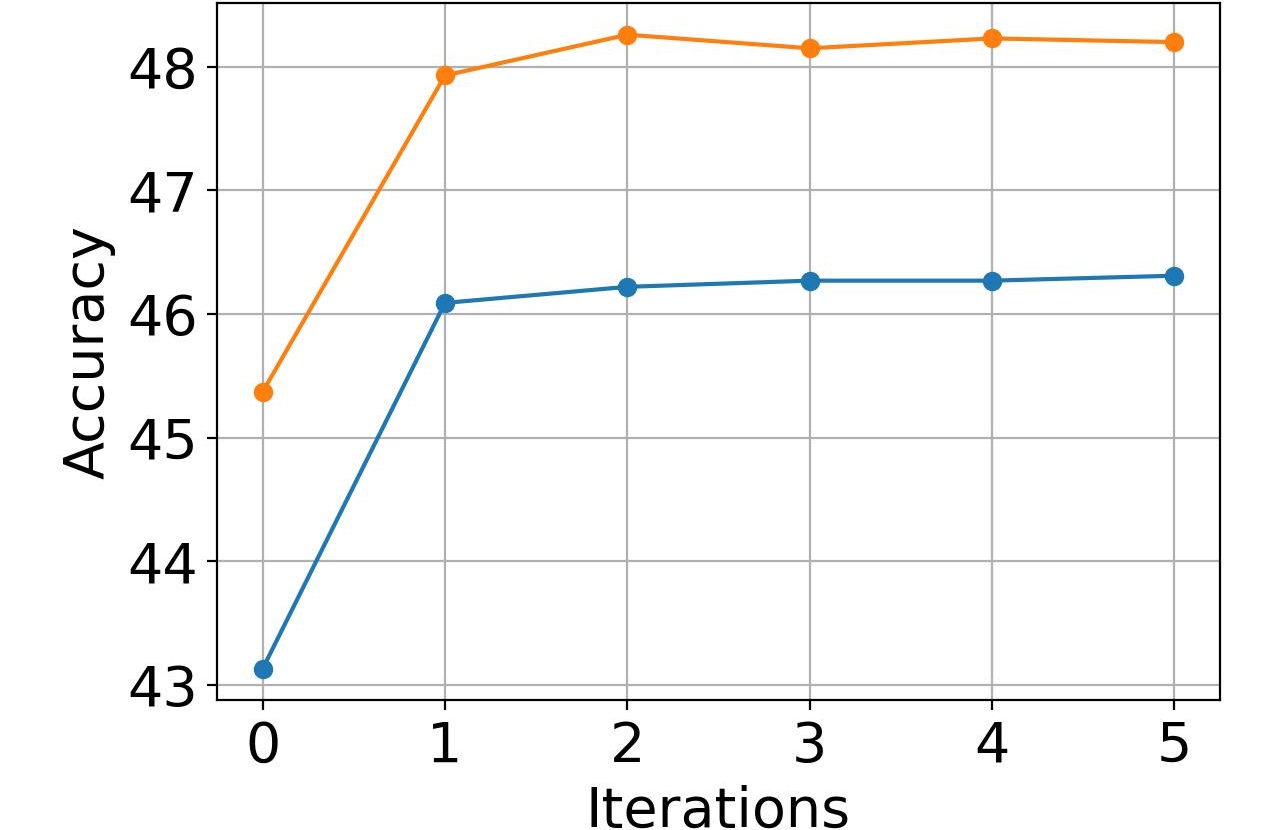}
  \caption{Performance on GQA.}
\end{subfigure}%
\begin{subfigure}{.33\textwidth}
  \centering
  \includegraphics[width=\textwidth]{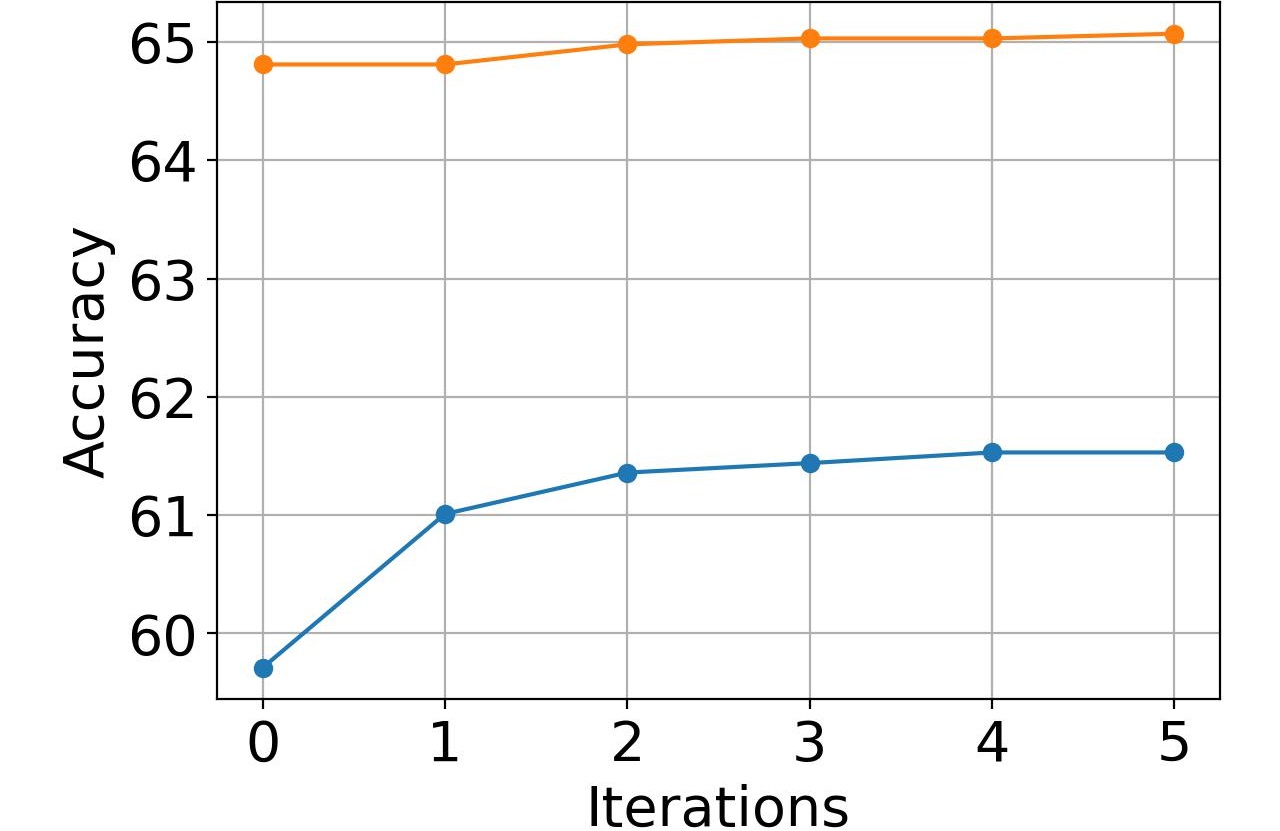}
  \caption{Performance on NLVR.}
\end{subfigure}%
\begin{subfigure}{.33\textwidth}
  \centering
  \includegraphics[width=\textwidth]{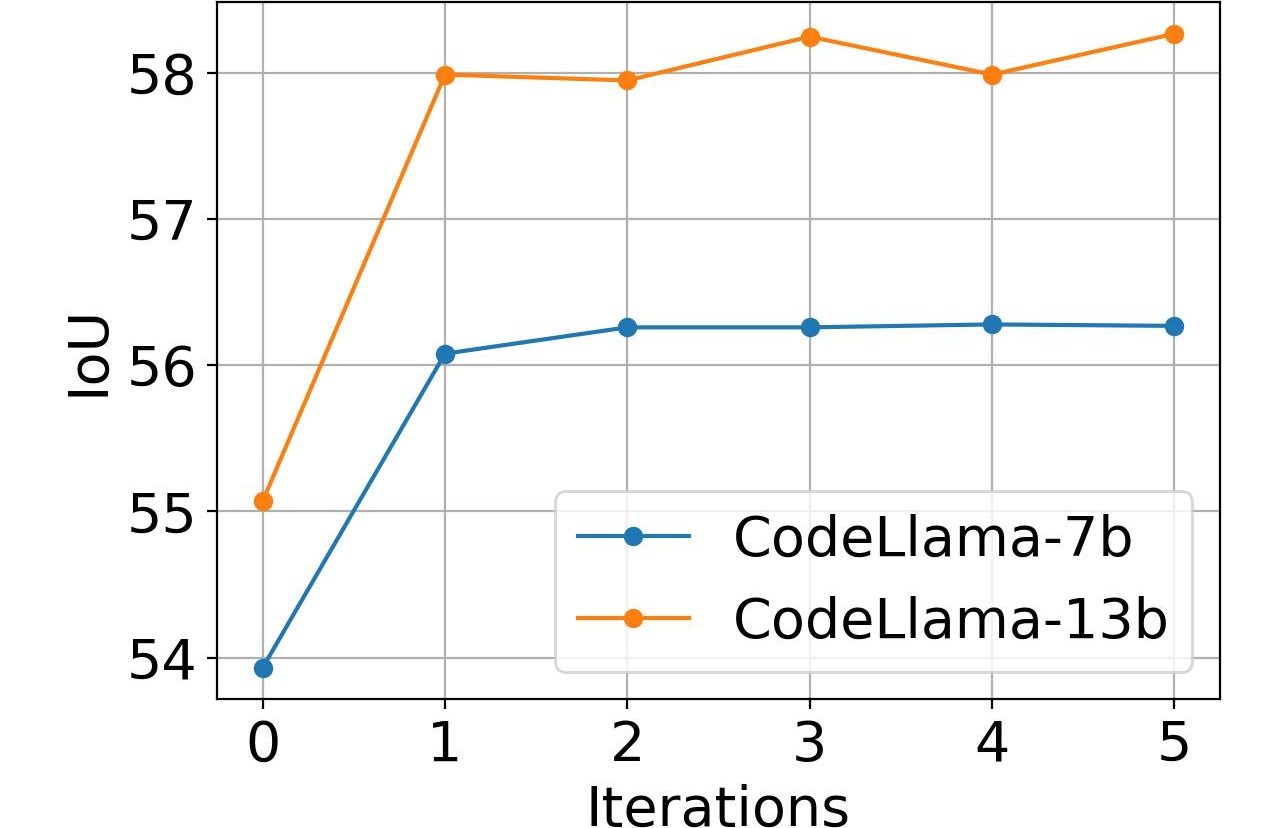}
  \caption{Performance on RefCOCOg.}
\end{subfigure}
\caption{Performance on GQA, NLVRv2 and RefCOCOg datasets by the number of debugging iterations.}
\label{fig:iteration}
\end{figure*}

\begin{table*}[!htbp]
\centering
\setlength{\tabcolsep}{4pt}
\resizebox{\textwidth}{!}{
\begin{tabular}{l|llllll|l}
\hline
& GQA & TallyQA & NLVRv2 & RefCOCO & RefCOCO+ & RefCOCOg & Mean \\
\hline
\multicolumn{5}{l}{\textit{CodeLlama-70b as code generator}} \\
\hline
No Debugging & 46.1 & 45.2 & 68.1 & 54.0 & 49.6 & 53.3 & 52.7 \\
\methodname{} 7b & 48.8 {\footnotesize \color{mygreen} (+2.7)} & 47.9 {\footnotesize \color{mygreen} (+2.7)} & \textbf{68.9} {\footnotesize \color{mygreen} (+0.8)} & \textbf{57.4} {\footnotesize \color{mygreen} (+3.4)} & \textbf{50.9} {\footnotesize \color{mygreen} (+1.3)} & \textbf{56.7} {\footnotesize \color{mygreen} (+3.4)} & \textbf{55.1} {\footnotesize \color{mygreen} (+2.4)} \\
\methodname{} 13b & \textbf{48.9} {\footnotesize \color{mygreen} (+2.8)} & \textbf{48.5} {\footnotesize \color{mygreen} (+3.3)} & 68.6 {\footnotesize \color{mygreen} (+0.5)} & 56.9 {\footnotesize \color{mygreen} (+2.9)} & \textbf{50.9} {\footnotesize \color{mygreen} (+1.3)} & 56.4 {\footnotesize \color{mygreen} (+3.1)} & 55.0 {\footnotesize \color{mygreen} (+2.3)} \\
\hline
\multicolumn{5}{l}{\textit{DeepSeek-Coder-33B as code generator}} \\
\hline
No Debugging & 47.3 & 46.2 & \textbf{67.9} & 59.6 & 51.8 & 55.4 & 54.7 \\
\methodname{} 7b & \textbf{48.9} {\footnotesize \color{mygreen} (+1.6)} & \textbf{46.9} {\footnotesize \color{mygreen} (+0.7)} & 67.8 {\footnotesize \color{red} (-0.1)} & \textbf{60.5} {\footnotesize \color{mygreen} (+0.9)} & 52.4 {\footnotesize \color{mygreen} (+0.6)} & 57.1 {\footnotesize \color{mygreen} (+1.7)} & \textbf{55.6} {\footnotesize \color{mygreen} (+1.9)} \\
\methodname{} 13b & \textbf{48.9} {\footnotesize \color{mygreen} (+1.6)} & 46.2 {\footnotesize \color{mygreen} (+0.0)} & 67.4 {\footnotesize \color{red} (-0.5)} & \textbf{60.5} {\footnotesize \color{mygreen} (+0.9)} & \textbf{52.8} {\footnotesize \color{mygreen} (+1.0)} & \textbf{57.7} {\footnotesize \color{mygreen} (+2.3)} & \textbf{55.6} {\footnotesize \color{mygreen} (+1.9)} \\
\hline
\multicolumn{5}{l}{\textit{GPT-3.5 as code generator}} \\
\hline
No Debugging & 45.9 & 43.6 & 63.9 & 60.6 & 54.0 & 57.5 & 54.3 \\
\methodname{} 7b & 49.3 {\footnotesize \color{mygreen} (+3.4)} & \textbf{46.4} {\footnotesize \color{mygreen} (+2.8)} & \textbf{68.8} {\footnotesize \color{mygreen} (+4.9)} & 61.2 {\footnotesize \color{mygreen} (+0.6)} & \textbf{54.7} {\footnotesize \color{mygreen} (+0.7)}  & \textbf{58.8} {\footnotesize \color{mygreen} (+1.3)} & \textbf{56.5} {\footnotesize \color{mygreen} (+2.2)} \\
\methodname{} 13b & \textbf{49.6} {\footnotesize \color{mygreen} (+3.7)} & \textbf{46.4} {\footnotesize \color{mygreen} (+2.8)} & 68.7 {\footnotesize \color{mygreen} (+4.8)} & \textbf{61.3} {\footnotesize \color{mygreen} (+0.7)} & 54.1 {\footnotesize \color{mygreen} (+0.6)} & \textbf{58.8} {\footnotesize \color{mygreen} (+1.3)} & \textbf{56.5} {\footnotesize \color{mygreen} (+2.2)} \\
\hline
\end{tabular}
}
\caption{\methodname{} can debug visual programs generated by larger LLMs, including CodeLlama-70b, DeepSeek-Coder-33B and GPT-3.5.}
\label{tab:even_stronger_codegen}
\end{table*}

\mypar{Ablation study.} We investigate the contribution of each component as shown in Table \ref{tab:main_ablation}. Specifically, we aim to: (1) assess the individual contributions of critic and refiner components, and (2) evaluate the benefits of execution feedback. We report the critic's binary accuracy in predicting overall program correctness, as well as the percentage of incorrect programs successfully fixed by refiner, denoted as refiner success rate. The critic demonstrates consistently strong performance, with high binary accuracy ranging from 67\% to 80\% across different datasets. Our manual evaluation of 59 examples from GQA shows the predicted error-inducing errors are correct in 74\% of the cases. However, the refiner success rate is less reliable, varying dramatically from 10\% to 57\% across datasets. When enhanced with execution feedback, the critic achieves more performance gains while the benefits to refiner performance are minimal. When reflected in the final performance on the downstream tasks, execution feedback consistently brings benefits on all datasets. In general, \methodname{} can reliably perform self-critique utilizing execution feedback, and the remaining challenges mainly lie in correcting the program after the errors are identified.

\mypar{Performance by iteration.} \methodname{} can perform iterative debugging until the critic determines the program as correct. Figure \ref{fig:iteration} demonstrates the performance curve by the number of iterations on three representative datasets for the three task forms, GQA, NLVRv2, and RefCOCOg. We find that most performance gains occur in the first one or two iterations, after which performance plateaus and may slightly decline. Qualitative analysis shows that more iterations are beneficial for complex problems, where the initial debugging attempt often fails, so \methodname{} need to iteratively refines the program in a trial-and-error manner. An example is shown in Figure \ref{fig:case_iteration} in the Appendix.

\mypar{Generalization to other code generators.}
\begin{table}[!htb]
\centering
\begin{tabular}{l|ll}
\hline
& RSVG & COVR \\
\hline
Base VLM & 18.1 & 41.2 \\
\hline
\multicolumn{3}{l}{\textit{CodeLlama-7b as code generator}} \\
\hline
No Debugging & 17.9 & 41.5 \\
\methodname{} & \textbf{18.7} {\footnotesize \color{mygreen} (+0.8)} & \textbf{43.8} {\footnotesize \color{mygreen} (+2.3)} \\
\hline
\multicolumn{3}{l}{\textit{CodeLlama-13b as code generator}} \\
\hline
No Debugging & 18.3 & 46.9 \\
\methodname{} & \textbf{18.8} {\footnotesize \color{mygreen} (+0.5)} & \textbf{47.9} {\footnotesize \color{mygreen} (+1.0)} \\
\hline
\end{tabular}
\caption{\methodname{} can generalize to unseen tasks, including visual grounding for remote sensing images (RSVG) and visual question answering over variable number of images (COVR). We report IoU for RSVG and accuracy for COVR.}
\label{tab:generalization}
\end{table}

While \methodname{} is trained on programs generated by CodeLlama models, it can be employed to debug programs generated by LLMs with larger number of parameters. As shown in Table~\ref{tab:even_stronger_codegen}, we experiment with two open LLMs, CodeLlama-70b and DeepSeek-Coder-33B \citep{deepseek-coder}, and the proprietary LLM GPT-3.5. Despite these models being up to ten times larger than our base models and achieving higher performance without any debugging, \methodname{}'s debugging process still consistently brings improvements, demonstrating its generalization capability. Thus, employing zero-shot large-scale LLMs debugged by a small \methodname{} can be a good strategy to enhance performance at a reasonable cost.

\mypar{Generalization to unseen tasks.} We evaluate the generalist variant VDebugger w/ Gen, which is trained on all six datasets, on two unseen datasets: (1) RSVG \citep{rsvg}, a visual grounding dataset for remote sensing images, a challenging task due to the dense objects and complex spatial relationships in remote sensing images; and (2) COVR \citep{chen-etal-2022-comparative-graph}, a novel task form requiring the model to answer questions based on a variable number of images. Table \ref{tab:generalization} shows that \methodname{} consistently improves performance on both datasets, demonstrating its ability to generalize to unseen domains and task formulations.

\begin{figure}[!htb]
    \centering
\includegraphics[width=\linewidth]{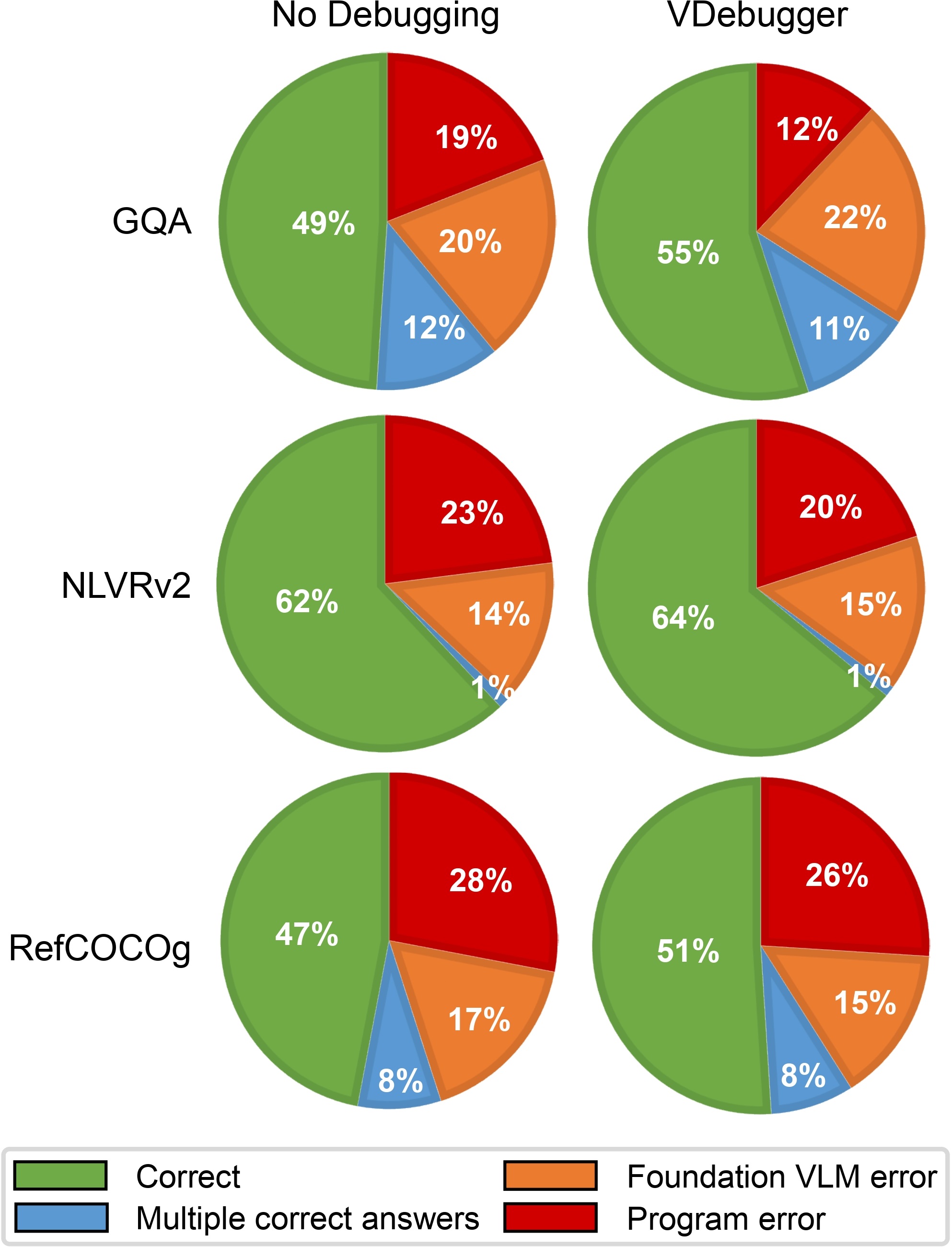}
    \caption{\small \textbf{Sources of errors on GQA, NLVRv2 and RefCOCOg datasets.} We categorize the predictions into four categories: correct, multiple correct answers (where the prediction is correct but does not match the ground truth annotation), foundation VLM errors, and program errors.}
    \label{fig:error}
\end{figure}

\mypar{Data quality.} To verify the quality of automatically generated data, we manually examine 100 programs from each training set. We evaluate the proportions of incorrect programs, or "false positives", among the programs considered correct. Most datasets have relatively low false positive ratio: 16\% for GQA, 13\% for TallyQA, and 19\% for RefCOCO. Due to the answer format, including free-form strings, numbers and bounding boxes, an exact match in the final answer ensures the program has a high probability to be correct. On the other hand, NLVRv2 dataset has a higher false positive ratio (40\%) due to its binary label format. However, its effect can be mitigated by training on multiple datasets, as shown by the generalist VDebugger outperforming the specialist VDebugger on NLVRv2 dataset. While another concern over data quality is the program optimality, we observe that visual programs tend to have straightforward code structures, and thus have limited potential for algorithmic optimization. For example, 68\% of the programs do not contain loop structures.

\begin{figure*}[!htb]
    \centering
\includegraphics[width=\linewidth]{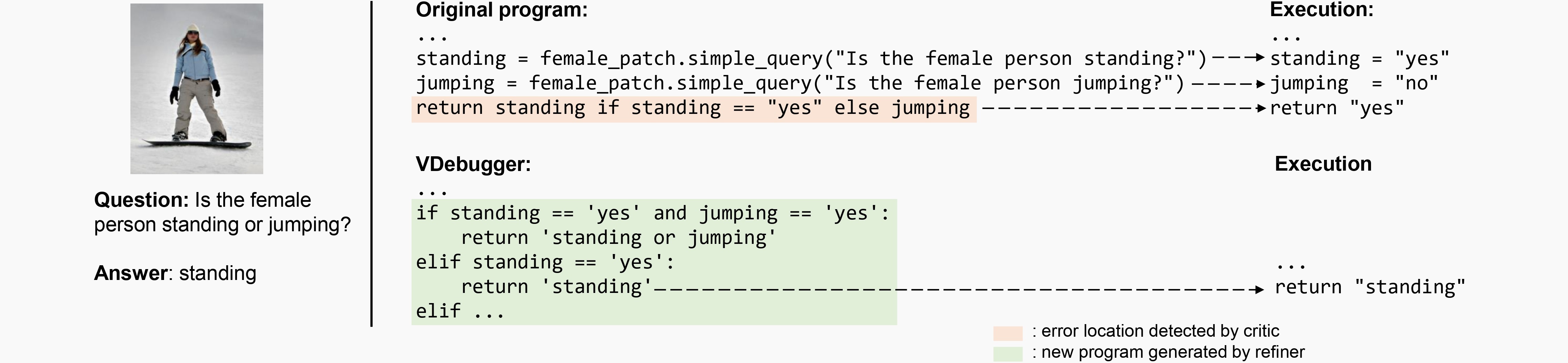}%
\vspace{-.4em}
    \caption{\small \textbf{Example where \methodname{} fixes program error.}}
    \label{fig:case_study}
\end{figure*}

\begin{figure*}[!htb]
    \centering
\includegraphics[width=\linewidth]{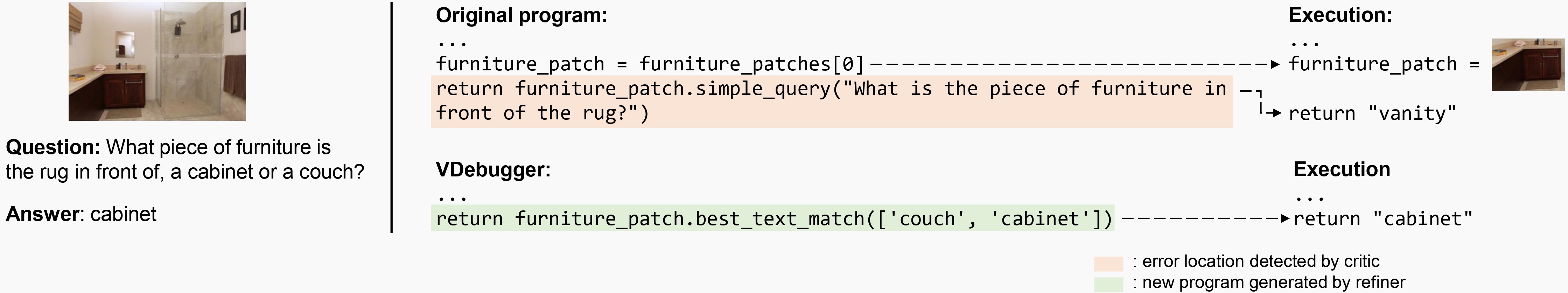}
\vspace{-1.4em}
    \caption{\small \textbf{Example where \methodname{} recovers from foundation model error.} The question answering model yields incorrect answer ``vanity'' in the original program. By detecting this error, \methodname{} invokes the foundation VLMs in an alternative way and thus obtains the correct answer.}
    \label{fig:case_study2}
\end{figure*}

\begin{table*}[!t]
    \centering
    \begin{minipage}{0.55\textwidth}
    \centering
\setlength{\tabcolsep}{4pt}
        \begin{tabular}{l|l|l|l}
            \hline
            & \multirow{2}{*}{\textbf{Cost (s/it)}} & \textbf{\small Compared with} & \textbf{\small Compared with} \vspace{-.4em} \\
            & & \textbf{\small No Debugging} & \textbf{\small last iteration} \\
            \hline
            T = 0 & 3.08 & - & - \\ %
            T = 1 & 4.44 & + 44\% & +44\% \\
            T = 2 & 4.72 & + 53\% & +6\% \\
            T = 3 & 4.81 & + 56\% & +2\% \\
            \hline
        \end{tabular}
    \end{minipage}
    \hfill
    \begin{minipage}{0.44\textwidth}
        \centering
        \begin{tabular}{l|l}
            \hline
            \textbf{Component} & \textbf{Cost (s/it)} \\
            \hline
Initial program generation & 1.42 \\
            Program execution & 1.66 \\
            Critic inference & 0.09 \\
            Refiner inference & 0.08 \\
            \hline
        \end{tabular}
    \end{minipage}
    \caption{\small \textbf{Computational cost of VDebugger}, measured by seconds per item (s/it). \textbf{Left:} Computational cost by iteration step $T$. $T=0$ represents the no debugging baseline. \textbf{Right:} Breakdown of the computational cost of each component.}
    \label{tab:computation}
\end{table*}

\mypar{Qualitative analysis.} We analyze the sources of errors by examining 100 examples from each of the three datasets: GQA, NLVRv2, and RefCOCOg. As shown in Figure \ref{fig:error}, program errors significantly affect the end performance, accounting for 49\% to 62\% of total errors varying by dataset. \methodname{} consistently reduces program errors on all datasets, especially on GQA. An example of \methodname{} fixing program error is in Figure \ref{fig:case_study}. Interestingly, we observe that \methodname{} can also help recover from foundation VLM errors especially on RefCOCOg dataset. While errors incurred by foundation VLMs remain a crucial bottleneck for visual programs, \methodname{} can invoke foundation VLMs in an alternative way to avoid the identified errors. An example is shown in Figure \ref{fig:case_study2}.

\mypar{Computational complexity.} To measure the additional computational overhead brought by the critic-refiner framework and the iterative process, we measure the computational cost by iteration step $T$ as well as a breakdown of the cost of each component in the framework. The detailed statistics are reported in Table \ref{tab:computation}. While VDebugger moderately increases the computational cost by 56\% compared to the no debugging baseline, the major computational overhead arises from program execution, rather than the inference of critic and refiner models. Additionally, increasing the number of debugging iterations only marginally increases the inference cost, since most debugging is addressed within the first round.

\section{Conclusion}

\methodname{} is a critic-refiner framework fine-tuned to detect, localize, and correct errors in visual programs leveraging fine-grained execution feedback. The training data is collected through an automated pipeline that first generates correct programs and then effectively injects errors using mask-best sampling. Experiments on six datasets demonstrate that \methodname{} consistently brings improvements, and further studies verifies \methodname{}'s generalization to unseen tasks. A future direction is to allow the visual program debugger to access visual information in addition to relying on textual information, and to jointly train it with foundation VLMs.

\section{Limitations}

We hereby discuss the potential limitations of our work:

(1) In this work, our critic model can provide basic explanations of identified errors by predicting errors locations. However, human programmers may benefit from more detailed explanations in natural language. The automatic collection of such text-rich description is very challenging. Therefore, obtaining expert annotations would be a valuable though costly future step to enhance the interpretability of the debugging process.

(2) Our work mainly focuses on established tasks such as visual question answering and visual grounding. While these tasks demonstrate the effectiveness of our framework, real-world applications often require systems to interact dynamically with humans, respond to open-ended questions, and perform on-demand reasoning. Although our current work does not directly address these complex, real-world scenarios, we believe our method is generic framework that can be adapted for such applications. Exploring the application of our self-debugging method to more in-the-wild and diverse scenarios is an exciting direction for future research.

(3) Following prior work \citep{visual_programming,vipergpt}, our method utilizes a text-only language model (LLM) to generate visual programs, which may introduce limitations to its capabilities. Incorporating visual information and/or jointly training the debugger with foundational VLMs could be a valuable direction for future research, potentially further enhancing its self-critic capabilities.

\section*{Acknowledgement}

This research is based upon work supported by CISCO, U.S. DARPA ECOLE Program No. \#HR00112390060, and OFFICE OF NAVAL RESEARCH Award \#N00014-23-1-2780. The views and conclusions contained herein are those of the authors and should not be interpreted as necessarily representing the official policies, either expressed or implied, of DARPA, or the U.S. Government. The U.S. Government is authorized to reproduce and distribute reprints for governmental purposes notwithstanding any copyright annotation therein.

\bibliography{anthology,custom}

\begin{thebibliography}{35}
\providecommand{\natexlab}[1]{#1}

\bibitem[{Acharya et~al.(2019)Acharya, Kafle, and Kanan}]{tallyqa}
Manoj Acharya, Kushal Kafle, and Christopher Kanan. 2019.
\newblock \href {https://doi.org/10.1609/AAAI.V33I01.33018076} {Tallyqa: Answering complex counting questions}.
\newblock In \emph{The Thirty-Third {AAAI} Conference on Artificial Intelligence, {AAAI} 2019, The Thirty-First Innovative Applications of Artificial Intelligence Conference, {IAAI} 2019, The Ninth {AAAI} Symposium on Educational Advances in Artificial Intelligence, {EAAI} 2019, Honolulu, Hawaii, USA, January 27 - February 1, 2019}, pages 8076--8084. {AAAI} Press.

\bibitem[{Bai et~al.(2023)Bai, Bai, Yang, Wang, Tan, Wang, Lin, Zhou, and Zhou}]{qwenvl}
Jinze Bai, Shuai Bai, Shusheng Yang, Shijie Wang, Sinan Tan, Peng Wang, Junyang Lin, Chang Zhou, and Jingren Zhou. 2023.
\newblock Qwen-vl: A versatile vision-language model for understanding, localization, text reading, and beyond.
\newblock \emph{arXiv preprint arXiv:2308.12966}.

\bibitem[{Bogin et~al.(2021)Bogin, Gupta, Gardner, and Berant}]{bogin-etal-2021-covr}
Ben Bogin, Shivanshu Gupta, Matt Gardner, and Jonathan Berant. 2021.
\newblock \href {https://doi.org/10.18653/v1/2021.emnlp-main.774} {{COVR}: A test-bed for visually grounded compositional generalization with real images}.
\newblock In \emph{Proceedings of the 2021 Conference on Empirical Methods in Natural Language Processing}, pages 9824--9846, Online and Punta Cana, Dominican Republic. Association for Computational Linguistics.

\bibitem[{Chen et~al.(2022)Chen, Cai, Jiang, and Chen}]{chen-etal-2022-comparative-graph}
Jingqiang Chen, Chaoxiang Cai, Xiaorui Jiang, and Kejia Chen. 2022.
\newblock \href {https://aclanthology.org/2022.coling-1.522} {Comparative graph-based summarization of scientific papers guided by comparative citations}.
\newblock In \emph{Proceedings of the 29th International Conference on Computational Linguistics}, pages 5978--5988, Gyeongju, Republic of Korea. International Committee on Computational Linguistics.

\bibitem[{Chen et~al.(2023)Chen, Lin, Schaerli, and Zhou}]{selfdebugging}
Xinyun Chen, Maxwell Lin, Nathanael Schaerli, and Denny Zhou. 2023.
\newblock Teaching large language models to self-debug.
\newblock In \emph{The 61st Annual Meeting Of The Association For Computational Linguistics}.

\bibitem[{Gao et~al.(2023)Gao, Li, Fei, Pang, Ji, Wang, Zhang, Tang, and Zhuang}]{define}
Minghe Gao, Juncheng Li, Hao Fei, Liang Pang, Wei Ji, Guoming Wang, Wenqiao Zhang, Siliang Tang, and Yueting Zhuang. 2023.
\newblock \href {https://doi.org/10.48550/ARXIV.2311.12890} {De-fine: Decomposing and refining visual programs with auto-feedback}.
\newblock \emph{CoRR}, abs/2311.12890.

\bibitem[{Guo et~al.(2024)Guo, Zhu, Yang, Xie, Dong, Zhang, Chen, Bi, Wu, Li, Luo, Xiong, and Liang}]{deepseek-coder}
Daya Guo, Qihao Zhu, Dejian Yang, Zhenda Xie, Kai Dong, Wentao Zhang, Guanting Chen, Xiao Bi, Y.~Wu, Y.~K. Li, Fuli Luo, Yingfei Xiong, and Wenfeng Liang. 2024.
\newblock \href {https://arxiv.org/abs/2401.14196} {Deepseek-coder: When the large language model meets programming -- the rise of code intelligence}.
\newblock \emph{Preprint}, arXiv:2401.14196.

\bibitem[{Gupta et~al.(2022)Gupta, Kamath, Kembhavi, and Hoiem}]{instructblip}
Tanmay Gupta, Amita Kamath, Aniruddha Kembhavi, and Derek Hoiem. 2022.
\newblock Towards general purpose vision systems: An end-to-end task-agnostic vision-language architecture.
\newblock In \emph{Proceedings of the IEEE/CVF Conference on Computer Vision and Pattern Recognition}, pages 16399--16409.

\bibitem[{Gupta and Kembhavi(2023)}]{visual_programming}
Tanmay Gupta and Aniruddha Kembhavi. 2023.
\newblock \href {https://doi.org/10.1109/CVPR52729.2023.01436} {Visual programming: Compositional visual reasoning without training}.
\newblock In \emph{{IEEE/CVF} Conference on Computer Vision and Pattern Recognition, {CVPR} 2023, Vancouver, BC, Canada, June 17-24, 2023}, pages 14953--14962. {IEEE}.

\bibitem[{Huang et~al.(2023)Huang, Chen, Mishra, Zheng, Yu, Song, and Zhou}]{DBLP:journals/corr/abs-2310-01798}
Jie Huang, Xinyun Chen, Swaroop Mishra, Huaixiu~Steven Zheng, Adams~Wei Yu, Xinying Song, and Denny Zhou. 2023.
\newblock \href {https://doi.org/10.48550/ARXIV.2310.01798} {Large language models cannot self-correct reasoning yet}.
\newblock \emph{CoRR}, abs/2310.01798.

\bibitem[{Hudson and Manning(2019)}]{gqa}
Drew~A. Hudson and Christopher~D. Manning. 2019.
\newblock \href {https://doi.org/10.1109/CVPR.2019.00686} {{GQA:} {A} new dataset for real-world visual reasoning and compositional question answering}.
\newblock In \emph{{IEEE} Conference on Computer Vision and Pattern Recognition, {CVPR} 2019, Long Beach, CA, USA, June 16-20, 2019}, pages 6700--6709. Computer Vision Foundation / {IEEE}.

\bibitem[{Jiang et~al.(2024)Jiang, Li, Wang, Zhou, Hossain, Ray, Kumar, Ma, and Deoras}]{jiang2024training}
Nan Jiang, Xiaopeng Li, Shiqi Wang, Qiang Zhou, Soneya~Binta Hossain, Baishakhi Ray, Varun Kumar, Xiaofei Ma, and Anoop Deoras. 2024.
\newblock \href {https://arxiv.org/abs/2405.18649} {Training llms to better self-debug and explain code}.
\newblock \emph{Preprint}, arXiv:2405.18649.

\bibitem[{Kamath et~al.(2023)Kamath, Hessel, and Chang}]{kamath2023s}
Amita Kamath, Jack Hessel, and Kai-Wei Chang. 2023.
\newblock What's" up" with vision-language models? investigating their struggle with spatial reasoning.
\newblock \emph{arXiv preprint arXiv:2310.19785}.

\bibitem[{Kamath et~al.(2024)Kamath, Hsieh, Chang, and Krishna}]{kamath2024hard}
Amita Kamath, Cheng-Yu Hsieh, Kai-Wei Chang, and Ranjay Krishna. 2024.
\newblock The hard positive truth about vision-language compositionality.
\newblock \emph{arXiv preprint arXiv:2409.17958}.

\bibitem[{Lan et~al.(2024)Lan, Zhang, Xu, Huang, Lin, Chen, and Mao}]{lan2024criticbench}
Tian Lan, Wenwei Zhang, Chen Xu, Heyan Huang, Dahua Lin, Kai Chen, and Xian-ling Mao. 2024.
\newblock Criticbench: Evaluating large language models as critic.
\newblock \emph{arXiv preprint arXiv:2402.13764}.

\bibitem[{Li et~al.(2023)Li, Li, Savarese, and Hoi}]{blip2}
Junnan Li, Dongxu Li, Silvio Savarese, and Steven C.~H. Hoi. 2023.
\newblock \href {https://proceedings.mlr.press/v202/li23q.html} {{BLIP-2:} bootstrapping language-image pre-training with frozen image encoders and large language models}.
\newblock In \emph{International Conference on Machine Learning, {ICML} 2023, 23-29 July 2023, Honolulu, Hawaii, {USA}}, volume 202 of \emph{Proceedings of Machine Learning Research}, pages 19730--19742. {PMLR}.

\bibitem[{Li et~al.(2022{\natexlab{a}})Li, Li, Xiong, and Hoi}]{blip}
Junnan Li, Dongxu Li, Caiming Xiong, and Steven C.~H. Hoi. 2022{\natexlab{a}}.
\newblock \href {https://proceedings.mlr.press/v162/li22n.html} {{BLIP:} bootstrapping language-image pre-training for unified vision-language understanding and generation}.
\newblock In \emph{International Conference on Machine Learning, {ICML} 2022, 17-23 July 2022, Baltimore, Maryland, {USA}}, volume 162 of \emph{Proceedings of Machine Learning Research}, pages 12888--12900. {PMLR}.

\bibitem[{Li et~al.(2022{\natexlab{b}})Li, Zhang, Zhang, Yang, Li, Zhong, Wang, Yuan, Zhang, Hwang, Chang, and Gao}]{glip}
Liunian~Harold Li, Pengchuan Zhang, Haotian Zhang, Jianwei Yang, Chunyuan Li, Yiwu Zhong, Lijuan Wang, Lu~Yuan, Lei Zhang, Jenq{-}Neng Hwang, Kai{-}Wei Chang, and Jianfeng Gao. 2022{\natexlab{b}}.
\newblock \href {https://doi.org/10.1109/CVPR52688.2022.01069} {Grounded language-image pre-training}.
\newblock In \emph{{IEEE/CVF} Conference on Computer Vision and Pattern Recognition, {CVPR} 2022, New Orleans, LA, USA, June 18-24, 2022}, pages 10955--10965. {IEEE}.

\bibitem[{Liu et~al.(2023)Liu, Li, Wu, and Lee}]{llava}
Haotian Liu, Chunyuan Li, Qingyang Wu, and Yong~Jae Lee. 2023.
\newblock Visual instruction tuning.
\newblock In \emph{NeurIPS}.

\bibitem[{Luo et~al.(2023)Luo, Lin, Liu, Shu, Zhu, Shang, and Meng}]{luo2023critique}
Liangchen Luo, Zi~Lin, Yinxiao Liu, Lei Shu, Yun Zhu, Jingbo Shang, and Lei Meng. 2023.
\newblock \href {https://arxiv.org/abs/2310.04815} {Critique ability of large language models}.
\newblock \emph{Preprint}, arXiv:2310.04815.

\bibitem[{Madaan et~al.(2023)Madaan, Tandon, Gupta, Hallinan, Gao, Wiegreffe, Alon, Dziri, Prabhumoye, Yang, Gupta, Majumder, Hermann, Welleck, Yazdanbakhsh, and Clark}]{selfrefine}
Aman Madaan, Niket Tandon, Prakhar Gupta, Skyler Hallinan, Luyu Gao, Sarah Wiegreffe, Uri Alon, Nouha Dziri, Shrimai Prabhumoye, Yiming Yang, Shashank Gupta, Bodhisattwa~Prasad Majumder, Katherine Hermann, Sean Welleck, Amir Yazdanbakhsh, and Peter Clark. 2023.
\newblock \href {http://papers.nips.cc/paper\_files/paper/2023/hash/91edff07232fb1b55a505a9e9f6c0ff3-Abstract-Conference.html} {Self-refine: Iterative refinement with self-feedback}.
\newblock In \emph{Advances in Neural Information Processing Systems 36: Annual Conference on Neural Information Processing Systems 2023, NeurIPS 2023, New Orleans, LA, USA, December 10 - 16, 2023}.

\bibitem[{Paul et~al.(2024)Paul, Ismayilzada, Peyrard, Borges, Bosselut, West, and Faltings}]{paul-etal-2024-refiner}
Debjit Paul, Mete Ismayilzada, Maxime Peyrard, Beatriz Borges, Antoine Bosselut, Robert West, and Boi Faltings. 2024.
\newblock \href {https://aclanthology.org/2024.eacl-long.67} {{REFINER}: Reasoning feedback on intermediate representations}.
\newblock In \emph{Proceedings of the 18th Conference of the European Chapter of the Association for Computational Linguistics (Volume 1: Long Papers)}, pages 1100--1126, St. Julian{'}s, Malta. Association for Computational Linguistics.

\bibitem[{Rachum et~al.(2019)Rachum, Hall, Yanokura et~al.}]{pysnooper}
Ram Rachum, Alex Hall, Iori Yanokura, et~al. 2019.
\newblock \href {https://doi.org/10.5281/zenodo.10462459} {Pysnooper: Never use print for debugging again}.

\bibitem[{Radford et~al.(2021)Radford, Kim, Hallacy, Ramesh, Goh, Agarwal, Sastry, Askell, Mishkin, Clark, Krueger, and Sutskever}]{clip}
Alec Radford, Jong~Wook Kim, Chris Hallacy, Aditya Ramesh, Gabriel Goh, Sandhini Agarwal, Girish Sastry, Amanda Askell, Pamela Mishkin, Jack Clark, Gretchen Krueger, and Ilya Sutskever. 2021.
\newblock \href {http://proceedings.mlr.press/v139/radford21a.html} {Learning transferable visual models from natural language supervision}.
\newblock In \emph{Proceedings of the 38th International Conference on Machine Learning, {ICML} 2021, 18-24 July 2021, Virtual Event}, volume 139 of \emph{Proceedings of Machine Learning Research}, pages 8748--8763. {PMLR}.

\bibitem[{Rozière et~al.(2024)Rozière, Gehring, Gloeckle, Sootla, Gat, Tan, Adi, Liu, Sauvestre, Remez, Rapin, Kozhevnikov, Evtimov, Bitton, Bhatt, Ferrer, Grattafiori, Xiong, Défossez, Copet, Azhar, Touvron, Martin, Usunier, Scialom, and Synnaeve}]{codellama}
Baptiste Rozière, Jonas Gehring, Fabian Gloeckle, Sten Sootla, Itai Gat, Xiaoqing~Ellen Tan, Yossi Adi, Jingyu Liu, Romain Sauvestre, Tal Remez, Jérémy Rapin, Artyom Kozhevnikov, Ivan Evtimov, Joanna Bitton, Manish Bhatt, Cristian~Canton Ferrer, Aaron Grattafiori, Wenhan Xiong, Alexandre Défossez, Jade Copet, Faisal Azhar, Hugo Touvron, Louis Martin, Nicolas Usunier, Thomas Scialom, and Gabriel Synnaeve. 2024.
\newblock \href {https://arxiv.org/abs/2308.12950} {Code llama: Open foundation models for code}.
\newblock \emph{Preprint}, arXiv:2308.12950.

\bibitem[{Shinn et~al.(2023)Shinn, Cassano, Gopinath, Narasimhan, and Yao}]{reflexion}
Noah Shinn, Federico Cassano, Ashwin Gopinath, Karthik Narasimhan, and Shunyu Yao. 2023.
\newblock \href {http://papers.nips.cc/paper\_files/paper/2023/hash/1b44b878bb782e6954cd888628510e90-Abstract-Conference.html} {Reflexion: language agents with verbal reinforcement learning}.
\newblock In \emph{Advances in Neural Information Processing Systems 36: Annual Conference on Neural Information Processing Systems 2023, NeurIPS 2023, New Orleans, LA, USA, December 10 - 16, 2023}.

\bibitem[{Stanic et~al.(2024)Stanic, Caelles, and Tschannen}]{towards_zero_shot_visual_programming}
Aleksandar Stanic, Sergi Caelles, and Michael Tschannen. 2024.
\newblock \href {https://doi.org/10.48550/ARXIV.2401.01974} {Towards truly zero-shot compositional visual reasoning with llms as programmers}.
\newblock \emph{CoRR}, abs/2401.01974.

\bibitem[{Suhr et~al.(2019)Suhr, Zhou, Zhang, Zhang, Bai, and Artzi}]{suhr-etal-2019-corpus}
Alane Suhr, Stephanie Zhou, Ally Zhang, Iris Zhang, Huajun Bai, and Yoav Artzi. 2019.
\newblock \href {https://doi.org/10.18653/v1/P19-1644} {A corpus for reasoning about natural language grounded in photographs}.
\newblock In \emph{Proceedings of the 57th Annual Meeting of the Association for Computational Linguistics}, pages 6418--6428, Florence, Italy. Association for Computational Linguistics.

\bibitem[{Sur{\'{\i}}s et~al.(2023)Sur{\'{\i}}s, Menon, and Vondrick}]{vipergpt}
D{\'{\i}}dac Sur{\'{\i}}s, Sachit Menon, and Carl Vondrick. 2023.
\newblock \href {https://doi.org/10.1109/ICCV51070.2023.01092} {Vipergpt: Visual inference via python execution for reasoning}.
\newblock In \emph{{IEEE/CVF} International Conference on Computer Vision, {ICCV} 2023, Paris, France, October 1-6, 2023}, pages 11854--11864. {IEEE}.

\bibitem[{Tian et~al.(2024)Tian, Ye, Qin, Cong, Lin, Pan, Wu, Liu, and Sun}]{tian2024debugbench}
Runchu Tian, Yining Ye, Yujia Qin, Xin Cong, Yankai Lin, Yinxu Pan, Yesai Wu, Zhiyuan Liu, and Maosong Sun. 2024.
\newblock \href {https://arxiv.org/abs/2401.04621} {Debugbench: Evaluating debugging capability of large language models}.
\newblock \emph{Preprint}, arXiv:2401.04621.

\bibitem[{Wolf et~al.(2020)Wolf, Debut, Sanh, Chaumond, Delangue, Moi, Cistac, Rault, Louf, Funtowicz, Davison, Shleifer, von Platen, Ma, Jernite, Plu, Xu, Le~Scao, Gugger, Drame, Lhoest, and Rush}]{wolf-etal-2020-transformers}
Thomas Wolf, Lysandre Debut, Victor Sanh, Julien Chaumond, Clement Delangue, Anthony Moi, Pierric Cistac, Tim Rault, Remi Louf, Morgan Funtowicz, Joe Davison, Sam Shleifer, Patrick von Platen, Clara Ma, Yacine Jernite, Julien Plu, Canwen Xu, Teven Le~Scao, Sylvain Gugger, Mariama Drame, Quentin Lhoest, and Alexander Rush. 2020.
\newblock \href {https://doi.org/10.18653/v1/2020.emnlp-demos.6} {Transformers: State-of-the-art natural language processing}.
\newblock In \emph{Proceedings of the 2020 Conference on Empirical Methods in Natural Language Processing: System Demonstrations}, pages 38--45, Online. Association for Computational Linguistics.

\bibitem[{Yu et~al.(2016)Yu, Poirson, Yang, Berg, and Berg}]{refcoco}
Licheng Yu, Patrick Poirson, Shan Yang, Alexander~C. Berg, and Tamara~L. Berg. 2016.
\newblock \href {https://doi.org/10.1007/978-3-319-46475-6\_5} {Modeling context in referring expressions}.
\newblock In \emph{Computer Vision - {ECCV} 2016 - 14th European Conference, Amsterdam, The Netherlands, October 11-14, 2016, Proceedings, Part {II}}, volume 9906 of \emph{Lecture Notes in Computer Science}, pages 69--85. Springer.

\bibitem[{Y{\"{u}}ksekg{\"{o}}n{\"{u}}l et~al.(2023)Y{\"{u}}ksekg{\"{o}}n{\"{u}}l, Bianchi, Kalluri, Jurafsky, and Zou}]{DBLP:conf/iclr/Yuksekgonul0KJ023}
Mert Y{\"{u}}ksekg{\"{o}}n{\"{u}}l, Federico Bianchi, Pratyusha Kalluri, Dan Jurafsky, and James Zou. 2023.
\newblock \href {https://openreview.net/pdf?id=KRLUvxh8uaX} {When and why vision-language models behave like bags-of-words, and what to do about it?}
\newblock In \emph{The Eleventh International Conference on Learning Representations, {ICLR} 2023, Kigali, Rwanda, May 1-5, 2023}. OpenReview.net.

\bibitem[{Zhan et~al.(2023)Zhan, Xiong, and Yuan}]{rsvg}
Yang Zhan, Zhitong Xiong, and Yuan Yuan. 2023.
\newblock \href {https://doi.org/10.1109/TGRS.2023.3250471} {Rsvg: Exploring data and models for visual grounding on remote sensing data}.
\newblock \emph{IEEE Transactions on Geoscience and Remote Sensing}, 61:1--13.

\bibitem[{Zhong et~al.(2024)Zhong, Wang, and Shang}]{ldb}
Lily Zhong, Zilong Wang, and Jingbo Shang. 2024.
\newblock \href {https://doi.org/10.48550/ARXIV.2402.16906} {{LDB:} {A} large language model debugger via verifying runtime execution step-by-step}.
\newblock \emph{CoRR}, abs/2402.16906.

\end{thebibliography}
\bibliographystyle{acl_natbib}

\appendix
\section{Artifacts}

This work involves the following artifacts:

\textbf{Datasets:} GQA \citep{gqa} distributed under CC-BY-4.0 license, TallyQA \citep{tallyqa} distributed under Apache-2.0 license license, NLVRv2 \citep{suhr-etal-2019-corpus} distributed under CC-BY-4.0 license, RefCOCO \citep{refcoco} (including RefCOCO, RefCOCO+ and RefCOCOg variants) distributed under Apache-2.0 license, RSVG \citep{rsvg} without license specified, and COVR \citep{bogin-etal-2021-covr} distributed under MIT license.

\textbf{Software:} We use transformers \citep{wolf-etal-2020-transformers} and deepspeed (\url{https://github.com/microsoft/DeepSpeed}) for model training, both distributed under Apache-2.0 license. We collect execution feedback of visual programs using \texttt{pysnooper} \citep{pysnooper} distributed under MIT license.

\textbf{Models:} We use CodeLlama \citep{codellama} distributed under Llama's own license\footnote{\url{https://github.com/meta-llama/llama/blob/main/LICENSE}} and DeepSeek-Coder \citep{deepseek-coder} distributed under MIT License.

~

\noindent This work creates the following artifacts:

\textbf{Datasets:} We collect training data for our \methodname{} based on GQA \citep{gqa}, TallyQA \citep{tallyqa}, NLVRv2 \citep{suhr-etal-2019-corpus} and RefCOCO \citep{refcoco} datasets. Detailed statistics are in Table \ref{tab:datasets}.

\textbf{Software:} The code for training and inference of \methodname{} and training data collection.

\textbf{Models:} The \methodname{} 7B and 13B models, trained on each individual dataset as well as the generalist model trained on all datasets.

~

In summary, all the artifacts involved permit research use. Our use is consistent with their intended use.
We plan to release our software, datasets and models with license Apache-2.0 license, which is compatible with the original access conditions. All our artifacts are limited to English and do not cover multilingual scenarios.

\section{Implementation Details of \methodname{}}

\definecolor{return}{HTML}{0087BD}
\definecolor{value}{HTML}{00A64F}
\definecolor{code}{HTML}{CC5500}

Since \methodname{} is implemented based on LLMs, we need to effectively represent execution feedback $\texttt{Execute}(P)$ and error location $loc$ with text. The execution feedback is tracked and formatted via \texttt{pysnooper} \citep{pysnooper}. An example is shown in Figure \ref{fig:trace}: the feedback representation covers {\color{return} the final return value}, {\textcolor{code}{each code line being executed}, {\color{value} their resulted change in intermediate variable values}, and \red{execution errors} if any.
To represent a local span $loc$ with text, instead of directly generating the starting and ending location, we represent it by repeating the original program and wrapping location $loc$ with some special tokens. An example is shown in \ref{fig:loc}.

\begin{figure*}[!htb]
~~~~\centering
\footnotesize
\begin{tabularx}{\textwidth}{X}
\hline
\textbf{\textit{Question:}} What item of furniture is not large? \\
\hline
\textbf{\textit{Program:}} \\
def execute\_command(image) -> str: \\
~~~~image\_patch = ImagePatch(image) \\
~~~~image\_patch = best\_image\_match(list\_patches=[ImagePatch(image)], content=['item'], return\_index=True) \\
~~~~return image\_patch.simple\_query('What item of furniture is not large?') \\
\hline
\textbf{\textit{Execution feedback:}} \\
\\
\textcolor{return}{-> None} \\
\\
\textcolor{code}{call~~~~~~~~~1 def execute\_command(image) -> str:} \\
\textcolor{code}{line~~~~~~~~~2~~~~~image\_patch = ImagePatch(image)} \\
\textcolor{return}{New var:....... image\_patch = ImagePatch(left=0, right=500, upper=375, lower=0, height=375, width=500, horizontal\_center=250.0, vertical\_center=187.5)} \\
\textcolor{code}{line~~~~~~~~~3~~~~~image\_patch = best\_image\_match(list\_patches=[ImagePatch(image)], content=['item'], return\_index=True)} \\
\textcolor{return}{Modified var:.. image\_patch = 0} \\
\textcolor{code}{line~~~~~~~~~4~~~~~return image\_patch.simple\_query('What item of furniture is not large?')} \\
\textcolor{red}{exception~~~~4~~~~~return image\_patch.simple\_query('What item of furniture is not large?')} \\
\textcolor{red}{Exception:..... AttributeError: 'int' object has no attribute 'simple\_query'} \\
\textcolor{red}{Call ended by exception} \\
\hline
\end{tabularx}
~~~~\caption{\textbf{Text representation of feedback information.} The feedback incorporates {\color{return} the final return value}, {\color{code} each code line being executed}, {\color{value} their resulted change in intermediate variable values}, and \red{execution errors} if any.}
~~~~\label{fig:trace}
\end{figure*}

\begin{figure*}[!htb]
~~~~\centering
\footnotesize
\begin{tabularx}{\textwidth}{X}
\hline
\textbf{$loc$:} return image\_patch.simple\_query('What item of furniture is not large?') \\
\hline
\textbf{\textit{Representation:}} \\
\\
def execute\_command(image) -> str: \\
~~~~image\_patch = ImagePatch(image) \\
~~~~image\_patch = best\_image\_match(list\_patches=[ImagePatch(image)], content=['item'], return\_index=True) \\
~~~~\texttt{$<$BUG$>$}\textcolor{code}{return image\_patch.simple\_query('What item of furniture is not large?')}\texttt{$<$BUG/$>$} \\
\hline
\end{tabularx}
~~~~\caption{\textbf{Text representation of location $loc$.} In this example, special tokens \texttt{$<$BUG$>$} and \texttt{$<$BUG/$>$} wraps {\color{code} the location of interests}.}
~~~~\label{fig:loc}
\end{figure*}

\section{Experimental Details}

\mypar{Base VLM:} We use the same set of base VLMs as in \citet{vipergpt}. To report the performance of base VLMs, we use the question answering model BLIP-2 \citep{blip2} for visual question answering tasks, and the object detection model GLIP \citep{glip} for visual grounding tasks. Since BLIP-2 can only take one image as input, we concatenate all images into one when handling multiple images, such as in the NLVRv2 and COVR datasets.

\mypar{\methodname{}:} For fine-tuning \methodname{}, we use CodeLlama-7B-Python and CodeLlama-13B-Python as the base model. We truncate the context length into within 1024 tokens. We use a total batch size of $128$ sentences per batch (including ), a learning rate of $2e-5$, a linear scheduler for learning rate, and a warmup ratio of $0.03$. We train the CodeLlama-7B-Python for 3 epochs and CodeLlama-13B-Python for 1 epoch on all datasets. With 4 A6000 GPU, the training of refiner takes $\sim$4 hours and the training of critic takes $\sim$12 hours. In inference, we use greedy decoding with $256$ as the maximum number of tokens.

\mypar{Evaluation:} We evaluate the models on the \texttt{testdev} split of GQA, the Test-Complex split of TallyQA, the \texttt{test1} split of NLVRv2, the \texttt{testA} split by UNC of RefCOCO and RefCOCO+, and the standard test set split by UMD for RefCOCOg. We report accuracy for GQA, TallyQA and NLVRv2, and IoU for RefCOCO, RefCOCO+ and RefCOCOg. For accuracy, following the setting of \citet{vipergpt}, we first preprocess the answer produced by our method by removing stopwords and then use exact matching.

\begin{table}[!t]
\centering
\begin{tabular}{l|rr}
\hline
& \textbf{BLIP} & \textbf{InstructBLIP} \\ \hline
End-to-end VLMs & 43.1 & 48.3 \\
Visual programming & 45.4 & 48.4 \\
VDebugger & \textbf{48.1} & \textbf{51.1} \\
\hline
\end{tabular}
\caption{Performance of end-to-end VLMs, vanilla visual programming approach \citep{vipergpt} without debugging, and our VDebugger evaluated on GQA dataset. We experiment with BLIP \citep{blip} following \citet{vipergpt} as well as the more powerful VLM InstructBLIP \citep{instructblip}.}
\label{tab:recent_vlm}
\end{table}

\section{Visual Programming v.s. End-to-End VLMs}

Visual programming and end-to-end VLMs are two different approaches to visual reasoning.
Visual programming invokes multiple foundation VLMs through code, while end-to-end VLMs directly take an image as input and generate texts as output. Despite their seemingly different methodologies, visual programming is a complementary technique that can be combined with end-to-end VLMs to offer additional benefits. Firstly, visual programming can integrate with more powerful VLMs to further enhance performance as shown in Table \ref{tab:recent_vlm}. Secondly, despite the rapid development of end-to-end VLMs, they still have difficulty reasoning with compositional concepts such as counting and spatial relationship. Visual programming offer benefits in tasks like such as compositional reasoning, counting, and enhancing interpretability.

\section{Qualitative Examples}

Figure \ref{fig:case_iteration} shows an example where more iterations of \methodname{} bring benefits.

\begin{figure*}[!htbp]
    \centering
\includegraphics[width=\linewidth]{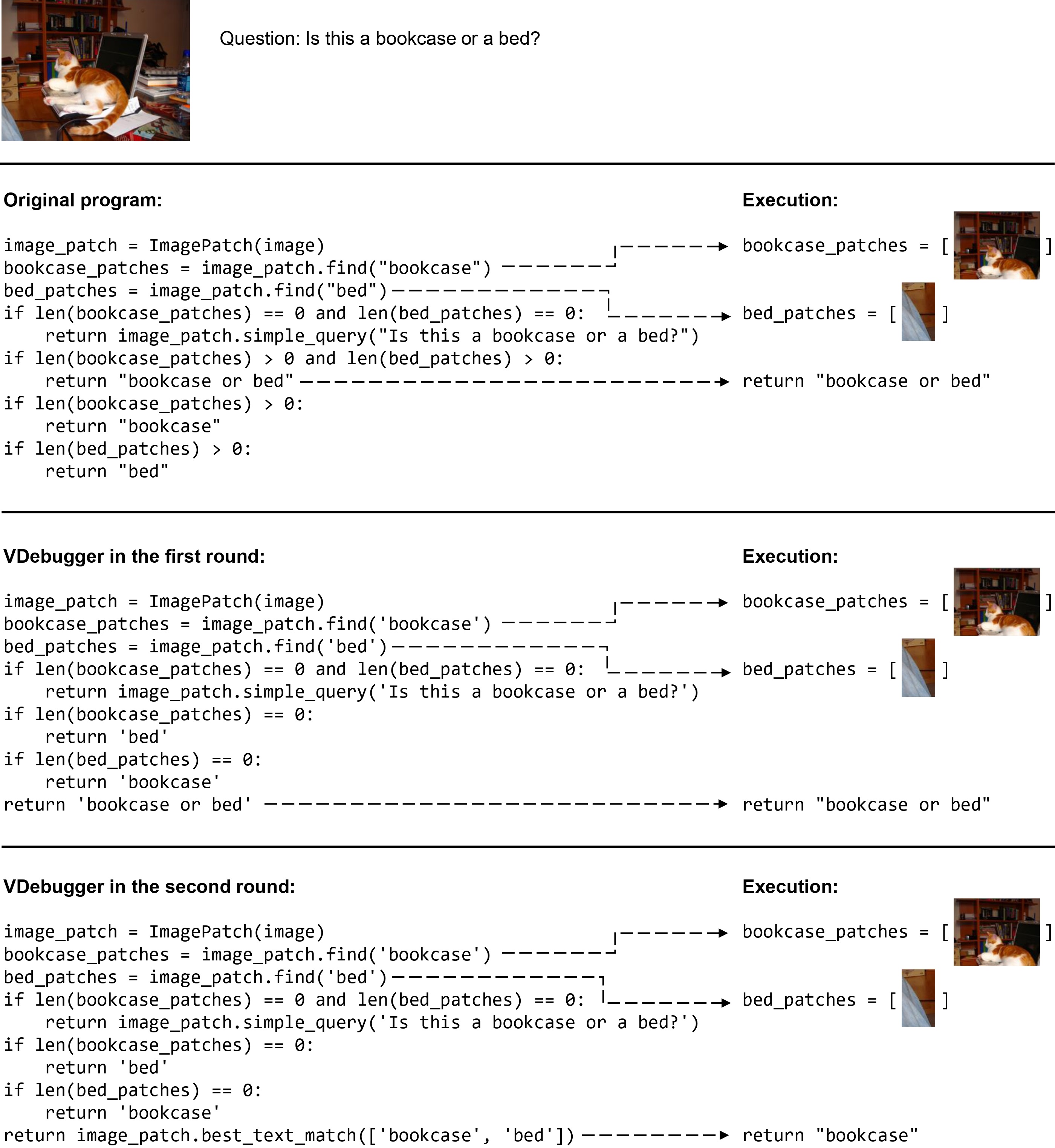}
    \caption{\textbf{Example where more iterations of \methodname{} bring benefits.} The original program results in incorrect answer ``bookcase or bed'' because the object detection model incorrectly identifies a bed. \methodname{} detects the error through the unreasonable return value and attempts the first round of debugging. Although the program structure is significantly changed in this round, the execution still leads to the incorrect answer due to the same issue. In the second round, \methodname{} successfully resolves the problem.}
    \label{fig:case_iteration}
\end{figure*}

\section{Prompts}

Figure \ref{fig:prompt_error_injection} shows the prompt we use for generating incorrect programs.

\begin{figure*}[!htbp]
~~~~\centering
\footnotesize
\begin{tabularx}{\textwidth}{X}
\hline
\textcolor{return}{[INST] I am writing code to handle visual question answering tasks by calling computer vision APIs. Some content from the code is masked (represented as "$<$MASKED$>$". Please recover the original code. }\\
\textcolor{return}{My code: }\\
\textcolor{return}{\textasciigrave\textasciigrave\textasciigrave python }\\
\textcolor{return}{\# \texttt{\{QUESTION\}} }\\
\textcolor{return}{\texttt{\{CODE\}} }\\
\textcolor{return}{\textasciigrave\textasciigrave\textasciigrave }\\
\textcolor{return}{}\\
\textcolor{return}{Your code should be wrapped in \textasciigrave\textasciigrave\textasciigrave python and \textasciigrave\textasciigrave\textasciigrave. The code should be exactly the same as my code, except recovering the masked content.}\\
\textcolor{return}{}\\
\textcolor{return}{---}\\
\textcolor{return}{}\\
\textcolor{return}{Below are the available APIs and some example usages:}\\
\textcolor{return}{}\\
\textcolor{return}{\textasciigrave\textasciigrave\textasciigrave python}\\
\textcolor{return}{\texttt{\{API\_DEFINITION\}}}\\
\textcolor{return}{\textasciigrave\textasciigrave\textasciigrave[/INST]} \textcolor{code}{Here's the original code with the \textasciigrave $<$MASKED$>$ \textasciigrave section replaced:}\\
\textcolor{code}{\textasciigrave\textasciigrave\textasciigrave python}\\
\textcolor{code}{\# \texttt{\{QUESTION\}}}\\
\textcolor{code}{\texttt{\{PROGRAM\_SIGNATURE\}}}\\
\hline
\end{tabularx}
~~~~\caption{Prompt for generating incorrect program. Here, {\color{code} the blue texts} are the prompt, {\color{code} the orange text} are the fixed prefix for model generation, and \texttt{\{QUESTION\}}, \texttt{\{CODE\}}, \texttt{\{API\_DEFINITION\}}, and \texttt{\{PROGRAM\_SIGNATURE\}} are placeholders to be filled in during actual generation.}
~~~~\label{fig:prompt_error_injection}
\end{figure*}

\end{document}